\documentclass{article}

\usepackage[accepted]{icmlcustom}

\usepackage[utf8]{inputenc} % allow utf-8 input
\usepackage{amsmath}
\usepackage[T1]{fontenc}    % use 8-bit T1 fonts
\usepackage{hyperref}       % hyperlinks
\usepackage{url}            % simple URL typesetting
\usepackage{booktabs}       % professional-quality tables
\usepackage{amsfonts}       % blackboard math symbols
\usepackage{nicefrac}       % compact symbols for 1/2, etc.
\usepackage{microtype}      % microtypography
\usepackage{cleveref}       % smart cross-referencing
\usepackage{graphicx}
\usepackage{subcaption}
\usepackage{xcolor}
\usepackage{bm}
\usepackage{dsfont}
\usepackage{multirow}
\usepackage{float}
\usepackage{tcolorbox}
\usepackage{enumitem}
\usepackage{rotating}

\usepackage[flushleft]{threeparttable}

\usepackage{pgfplots}
\icmltitlerunning{Efficient Embedding Adaptation using Contrastive Learning}

\begin{document}

\twocolumn[
\icmltitle{Efficient Domain Adaptation of Multimodal embeddings\\using Constrastive Learning}

% Set author information
\begin{icmlauthorlist}
\icmlauthor{Georgios Margaritis}{mit}
\icmlauthor{Periklis Petridis}{mit}
\icmlauthor{Dimitris J. Bertsimas}{sloan}
\end{icmlauthorlist}

\icmlaffiliation{mit}{Operations Research Center, Massachusetts Institute of Technology}
\icmlaffiliation{sloan}{Sloan School of Management, Massachusetts Institute of Technology}

\icmlcorrespondingauthor{Georgios Margaritis}{geomar@mit.edu}
\icmlcorrespondingauthor{Periklis Petridis}{periklis@mit.edu}
%\icmlcorrespondingauthor{Dimitris J. Bertsimas}{dbertsim@mit.edu}

\icmlkeywords{Contrastive Learning, NLP, Clinical Notes}

\vskip 0.3in
]

% Print author affiliations and notices
\printAffiliationsAndNotice{}

\begin{abstract}

     Recent advancements in machine learning (ML), natural language processing (NLP), and foundational models have shown promise for real-life applications in critical, albeit compute-constrainted fields like healthcare.  
     In such areas, combining foundational models with supervised ML offers potential for automating tasks like diagnosis and treatment planning, but the limited availability of onsite computational resources pose significant challenges before applying these technologies effectively: Current approaches either yield subpar results when using pretrained models without task-specific adaptation, or require substantial computational resources for fine-tuning, which is often a barrier to entry in such environments. 
     This renders them inaccessible in applications where performance and quality standards are high, but computational resources are scarce. 
     To bridge the gap between best-in-class performance and accessibility, we propose a novel method for adapting foundational, multimodal embeddings to downstream tasks, without the need of expensive fine-tuning processes. 
     Our method leverages frozen embeddings from Large Language Models (LLMs) and Vision Models, and uses contrastive learning to train a small, task-specific nonlinear projection that can be used in the downstream task, without having to fine-tune the original foundational models. 
     We show that this efficient procedure leads to significant performance improvements across various downstream tasks, and perhaps more importantly with minimal computational overhead, offering a practical solution for the use of advanced, foundational ML models in resource-constrained settings.
\end{abstract}

\section{Introduction}

%\lipsum[2]
%\lipsum[3]

Machine learning (ML) has grown very rapidly over the last 2 decades. This growth has inadvertently shown promise for its application in critical real-life applications, such as in clinical settings and healthcare operations. 
Although adoption of machine learning techniques in safety critical applications has had some inertia and initial skepticism, recent progress on safety, fairness, interpretability of ML models, as well as their performance, has created interest and momentum for their adoption from domain experts and practitioners.
Now, with the advent of rapid progress in foundational models (FMs) and the demonstration of accessible, general purpose capabilities from recent large language models (LLMs) \citep{bubeck2023sparks, brown2020language}, the interest from practitioners has piqued. However, significant challenges still remain for for the actual adoption of such models. 

On the one hand, FMs like Llama \citep{llama3}, BERT \cite{bert}, CLIP \citep{radford_learning_2021}, DINOv2 \citep{oquab2023dinov2}, etc. can produce high quality embeddings from multiple modalities which offer great performance on complex downstream tasks.
Hence, these models can be used out-of-the-box as feature extractors for various data types without the need for task-specific adaptation or fine-tuning \cite{haim}. 
However, this approach often yields suboptimal results, as the resulting embeddings may be too general for specific tasks and frequently high-dimensional, potentially introducing noise in the classifier. Such performance limitations are particularly concerning in safety-critical applications like healthcare, where accuracy and reliability are necessary for real-world deployment.

On the other hand, fine-tuning these foundation models to adapt to specific downstream tasks is computationally demanding:
It typically requires significant resources, such as large GPU servers and dedicated personnel, which are often missing from practical settings such as hospitals. 
To make matters worse, hospitals in particular, have to rely on the limited on-site computational resources,
since sharing their sensitive data with third-party cloud providers is not an option.
Hence, the cost and complexity of fine-tuning models with billions of parameters make it impractical 
or infeasible for most such real-world applications.
% The cost and complexity of fine-tuning models with billions of parameters make it impractical or even infeasible for most real-world applications, especially in resource-constrained environments.

As it stands currently, adoption of such models for practical assistance hinges on either (i) achieving very high model performance to satisfy the high standards of practitioners, which is often not feasible when using models without task-specific adaptation, or (ii) practitioners committing significant resources to build the necessary infrastructure to perform task-specific adaptation of the FMs. While this level of investment may occur in some instances, it severely limits the widespread adoption of such methods, especially in resource-constrained environments like many healthcare settings. This limitation creates a substantial barrier to leveraging the full potential of foundation models in critical real-world applications where they could provide the most benefit. 

To tackle this problem, our approach offers an efficient way to adapt embeddings to a downstream task, without requiring the fine-tuning of foundation models. 
We propose an inexpensive, task-specific embedding adaptation process using contrastive learning which employs a nonlinear projection of the original embeddings into a lower-dimensional space, trained to bring embeddings with the same labels closer together while separating those with different labels. 
Our experiments on real-world clinical notes demonstrate significant improvements in performance, with increases in test F1 score of up to more than 20\%, depending on the downstream model used. 
This approach provides a practical solution for leveraging the power of advanced ML models in resource-constrained settings, potentially facilitating wider adoption of these technologies in critical fields like healthcare.

Most contrastive learning methods require updating the entire embedding model in order to improve the quality of the embeddings. This process is computationally expensive, as it requires backpropagating the loss and updating the entire model, which can be costly both in terms of resources and time. Furthermore, many approaches also utilize implicit relationships between modalities, such as images and their captions, which may not be present in all multimodal settings. 
Instead, in our work, we develop a low-resource method that adapts to the task at hand efficiently, and only utilize the larger foundational models exactly once - to extract the initial embeddings.  After this point, we train a small non-linear model to map the original embeddings to a new task-specific subspace using a contrastive objective, improving the quality of the embeddings for the task at hand. 
Then, we can use the transformed embedding for our downstream task.\par 
%We can then use the transformed embeddings for our task. 
Throughout this process, it is important to reiterate that the foundational model is not updated, and therefore the initial embeddings do not change as we train or modify our pipeline. This means we can run each of these large models for inference once on each modality, extract the embeddings, and then simply store them for all future usage. As a matter of fact, the embeddings are potentially cheaper to store than the original raw data. Even if we modify our pipeline, adding or combining new modalities, or even if we want to adapt to a new task, we will not need to re-run the foundational models. After extracting the initial embeddings, our pipeline can run on a common CPU in less than a minute for thousands of samples.

To summarize, while our approach builds upon foundational contrastive learning methods, our contributions are the following:

\begin{enumerate}

    \item We introduce a novel and computationally efficient method
    for task-specific adaptation of multimodal embeddings, which significantly improves embedding performance in downstream classification tasks.
    Our method uses contrastive learning to train a small nonlinear projection on top of frozen embeddings, and doesn't require
    fine-tuning of entire embedding models.
    %, thus making it very computationally efficient.

    %\item Our method 
    \item We test the effectiveness of our method for diagnosing health conditions from real-world clinical notes. 
    Our experiments show that in most cases, our task-specific embeddings significantly outperform the original embeddings, increasing test F1 score by up to $20\%$ or more, depending on the feature extractor and downstream ML model used. 
    We also demonstrate that our results generalize to multiple ML models, multiple feature extractors and different multimodal datasets.
    
    \item We show that our method has very low computational requirements, and can run in minutes in CPU-only settings.
    % In fact, has a very similar execution time as the baseline
    It thus makes the usage of foundational models more accessible in resource-constrained environments, such as healthcare settings.
    Additionally, our method can also reduce computational requirements by enabling label-aware dimensionality reduction, with performance significantly better than PCA.
        
    \item Unlike conventional contrastive learning approaches tied to specific input types or implicit relationships (e.g. images and their captions), our method is modality-agnostic, since it operates on embeddings rather than raw inputs.
    This enables seamless integration of additional modalities using general-purpose embedding models, making it applicable across diverse data types and domains.

\end{enumerate}

These contributions collectively offer a simple, efficient, and flexible approach to leveraging powerful foundation models in resource-constrained environments, addressing specific challenges in settings like healthcare where computational resources may be limited, data appears in various forms, and task-specific adaptation is crucial for accurate predictions.

% Single column version
% \begin{figure}[ht]
%     \vskip 0.2in
%     \begin{center}
%     \centerline{\includegraphics[width=\columnwidth]{figs/pip1.pdf}} %{figs/pipeline_before.png}}
%     \caption{Multimodal prediction with task-agnostic embeddings}
%     \label{fig:pipeline1}
%     \end{center}
%     \vskip -0.2in
% \end{figure}

% Full width version (comment above and uncomment below)
% \begin{figure*}[t]
%     \vskip 0.2in
%     \begin{center}
%     \centerline{\includegraphics[width=0.85\textwidth]{figs/pipeline_before.png}}
%     \caption{Multimodal prediction with task-agnostic embeddings}
%     \label{fig:pipeline1}
%     \end{center}
%     \vskip -0.2in
% \end{figure*}

\section{Related Work}
\cite{gutmann_noise-contrastive_2010} laid out the groundwork for what is now contrastive learning by introducing Noise Contrastive Estimation (NCE) to estimate probabilistic models without computing partition functions. 
Word2Vec \citep{mikolov2013efficient} was one of the early applications of NCE in Natural Language Processing (NLP) and also introduced negative sampling to train word embeddings efficiently.
More recent developments using contrastive learning have led to significant advancements in representation learning. 

\textbf{Contrastive Learning in Vision.}
Notably, most of the recent work in this area has focused on unsupervised visual representation learning by utilizing data augmentations.
For instance, SimCLR \citep{chen_simple_2020} utilizes contrastive learning with data augmentation to learn visual representations. 
SwAV~\cite{caron_unsupervised_2020} combines clustering with contrastive learning for unsupervised visual representation learning. 
MoCo~\cite{he_momentum_2020} introduces a momentum contrast mechanism for unsupervised visual representation learning. 
In contrast, our method focuses on supervised tasks and uses task labels to adapt embeddings to the downstream task.

\textbf{Contrastive Learning in Multimodal Settings.}
In multimodal settings, contrastive learning has been used to learn joint representations of different modalities, typically images and text, by utilizing an implicit relationship, say images and their captions.
One of the first major works is Contrastive Predictive Coding (CPC)~\cite{oord_representation_2019} which introduces a way to learn representations by predicting future observations in latent space. 
Another notable work is CLIP~\cite{radford2021learning} which learns joint representations of images and text by pre-training on a large dataset of image-caption pairs. Other examples include ConVIRT~\cite{zhang_contrastive_2022} which applies contrastive learning to medical imaging by leveraging text reports associated with images and ALIGN~\cite{jia_scaling_2021}, which scales up visual and vision-language representation learning with noisy text supervision. However, these methods are designed for specific input types and rely on an implicit relationship between modalities, which may not be present in all multimodal settings. Our method is modality-agnostic and can be applied to any modality or combination thereof, as long as we have a general-purpose embedding model that can process that modality.

\textbf{Contrastive Learning in Clinical Settings.}
A ripe area for application of contrastive learning techniques is in clinical settings, where we frequently have multimodal data.
To this end, ConVIRT~\cite{zhang_contrastive_2022} applies contrastive learning to medical imaging by leveraging radiology reports associated with x-ray images.
A different approach is \cite{xiacontrastive}, which utilizes contrastive learning to improve representations of MLP and encoder architectures with partially missing sources for patients' tabular data.
However, the above methods are again constrained by the structure of the data or its implicit relationships. Our approach can utilize any type of data, and can be applied to any binary classification task.

\section{Methodology}
In this section we outline how we implement Contrastive Learning to perform task-specific embedding adaptation.
We first outline the problem setting, and then we analyze our Contrastive Learning approach.
\subsection{Problem Setting}
Consider the problem of binary classification based on multimodal data, where we have data from $m$ different sources/modalities.
\begin{figure}[ht]
    \vskip 0.2in
    \begin{center}
    \centerline{\includegraphics[width=\columnwidth]{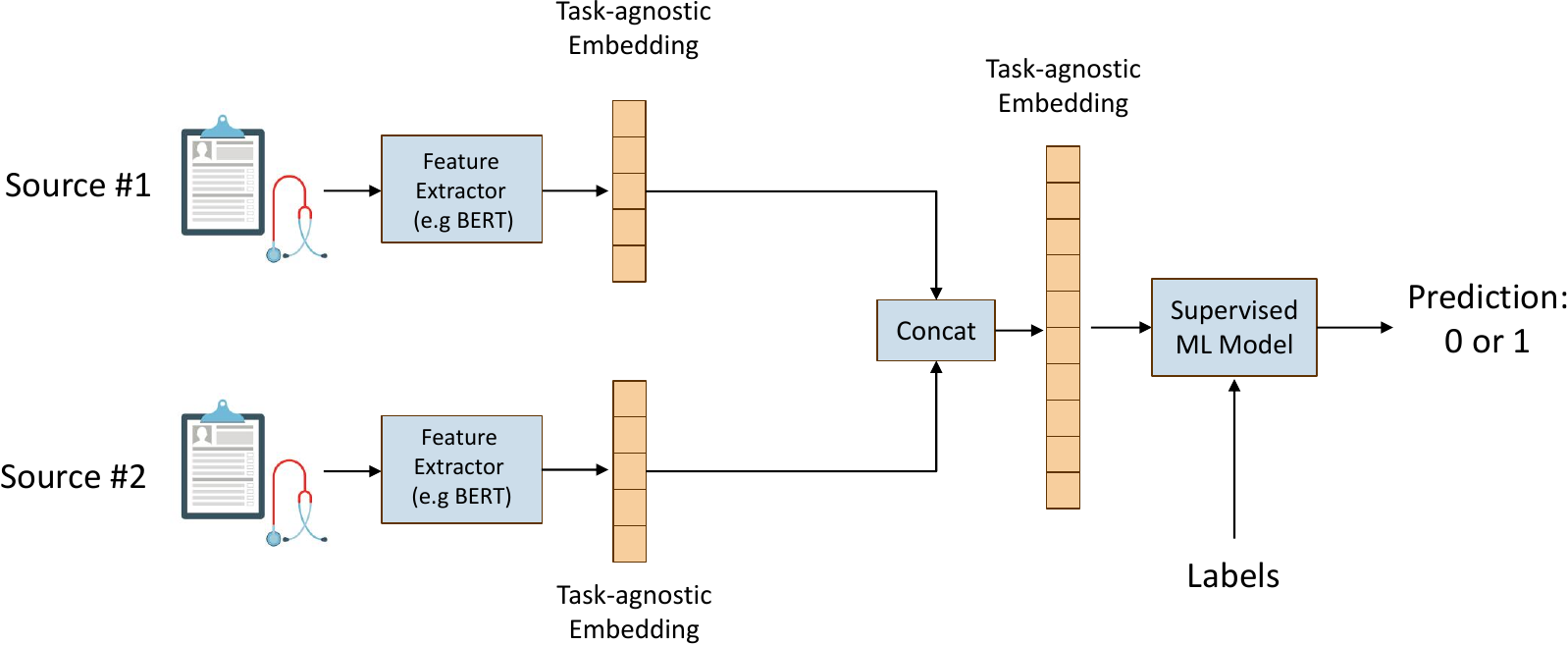}} %{figs/pipeline_before.png}}
    \caption{Multimodal prediction with task-agnostic embeddings}
    \label{fig:pipeline1}
    \end{center}
    \vskip -0.2in
\end{figure}

A typical approach to address this problem is to use a feature extractor to generate features/embeddings from every source, and then fuse all the features and then train a supervised ML model to predict the desired class from the fused features. 
% For clarity, we will refer to the output generated from the feature extractor as \emph{source embeddings}, and the fused features as \emph{task embeddings}. 
For instance, in the clinical setting, we may have textual data from clinical notes, images from radiology scans, and structured data from lab results, and we want to predict a binary label (e.g. disease vs. no disease) based on these multimodal data sources.
This pipeline is illustrated in Figure (\ref{fig:pipeline1}) and has been successfully applied in large-scale multimodal clinical data, as shown in \cite{haim}.\par

In this setting, each source embedding can be generated using a different feature extractor, depending on the nature of the data source.
For instance, extracting embeddings from textual sources can be done with BERT \cite{bert}, Llama3 combined with LLM2Vec \citep{llama3,llm2vec} or Qwen2 combined with GTE \citep{qwen2, gte}.
For visual sources, we can use models like DINOv2 \cite{dino}, Vision Transformers (ViT) \citep{dosovitskiy2020image}, CLIP \citep{radford_learning_2021}, and others. \par 

In all of these cases, we can find publicly-available versions of these models that we can use to extract embeddings without training these architectures from scratch.
This is particularly useful in scenarios where we have limited computational resources and we cannot resort to expensive task-specific training or finetuning procedures (such as using Masked Language Modeling to adapt large-scale LLMs to our task).
For instance, in most real-world applications, training, or even fine-tuning an LLM with billions of parameters to a specific task can be extremely costly, impractical or even infeasible. \par 
% However, embedding-generation without any form of task-adaptation can have various drawbacks. 
However, task-agnostic embedding generation can have serious drwabacks.
Using pre-trained feature extractors without any form of finetuning to the downstream task produces general-purpose and task-agnostic embeddings.
These embeddings may perform poorly in the specific downstream task.
At the same time, such embeddings can also be very high-dimensional (e.g. 4096 for Llama3), which may introduce a lot of noise in the resulting classifier.\par 

To mitigate both of these issues, we propose an inexpensive way of
creating task-specific embeddings, which in turn demonstrate significantly better performance than the task-agnostic embeddings in the binary classification task.
Our method uses an inexpensive nonlinear projection of the original embeddings into a lower-dimensional space.
This nonlinear projection is trained using Constrastive Learning, such that the output vectors of the projection that correspond to the same binary labels are close to one another, and the output vectors that correspond to different labels are far from one another.

% To mitigate both of those issues, we propose an inexpensive, task-specific embedding adaptation process which leverages contrastive learning to create low-dimensional task-specific embeddings

%significantly improve performance in the downstream task. 
% Our method uses an inexpensive nonlinear projection of the original embeddings into a lower-dimensional space.
% This nonlinear projection is trained using Constrastive Learning, such that the output vectors of the projection that correspond to the same labels are close to one another, and the output vectors that correspond to different labels are far from one another.\par 

% We will test our approach in the clinical setting, but we note that this approach is very general and is directly generalizable to other domains.
%We test this approach both on a private, clinical dataset and on a publicly available multimodal dataset.

% We first outline the standard prediction pipeline that is widely 
% used but doesn't incorporate embedding adaptation.
% Then, we explain how we modify this pipeline to adapt the feature vectors/embeddings to our downstream tasks via contrastive learning.

\subsection{Approach}
\label{sec:approach}
In this section, we describe our approach for task-specific embedding adaptation.
% Namely, instead of concatenating pre-trained embeddings and using them directly in a downstream classification task,
% we first adapt the embeddings to the task and then we use the adapted embeddings in downstream classifiers.
% This adaptation is done through lightweight nonlinear projections, which are trained through contrastive learning.

Assume we have $n$ data points and for each data point $i\in [n]$ we have task-agnostic embeddings $\bm{e}_i^{(1)}, \dots, \bm{e}_i^{(m)}$, where $m$ corresponds to the number of modalities.
For each data point, we also have a label $y_i$ which we assume to be binary.

\begin{figure}[ht]
    \vskip 0.2in
    \begin{center}
    \begin{subfigure}{\columnwidth}
        \centerline{\includegraphics[width=\columnwidth]{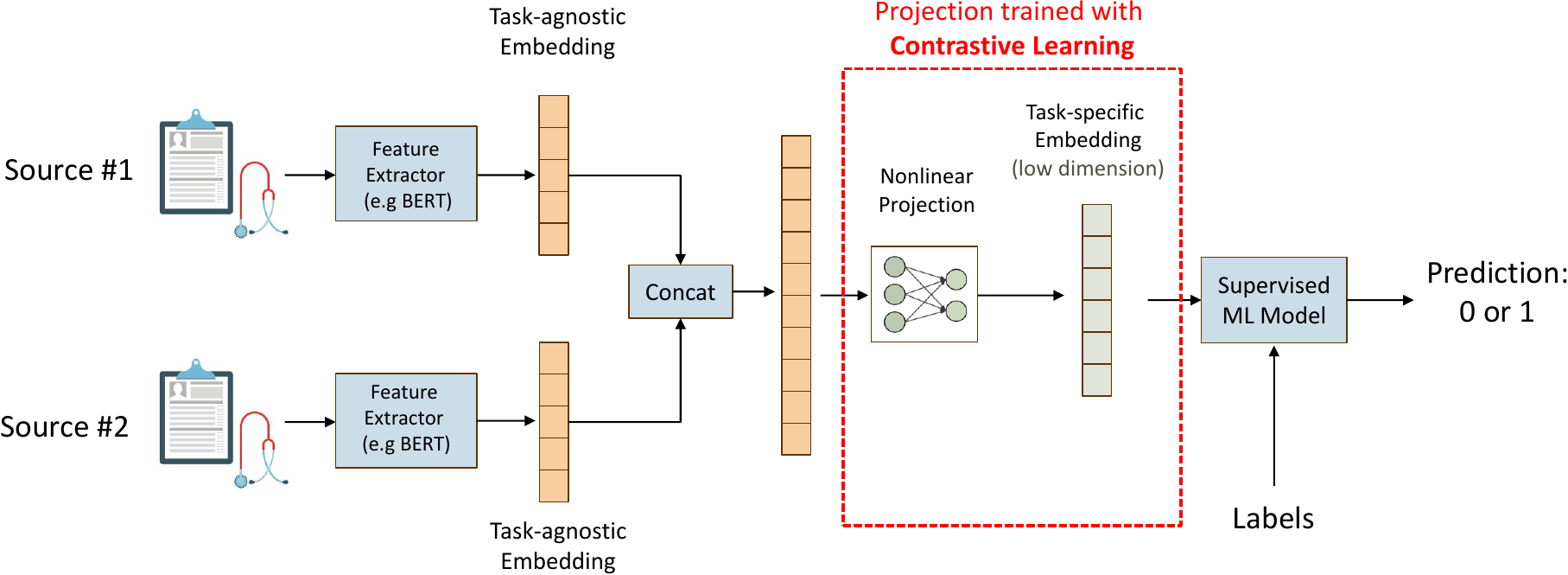}}
        \caption{Single Projection}
        \label{fig:single-proj}
    \end{subfigure}
    \vskip 0.1in
    \begin{subfigure}{\columnwidth}
        \centerline{\includegraphics[width=\columnwidth]{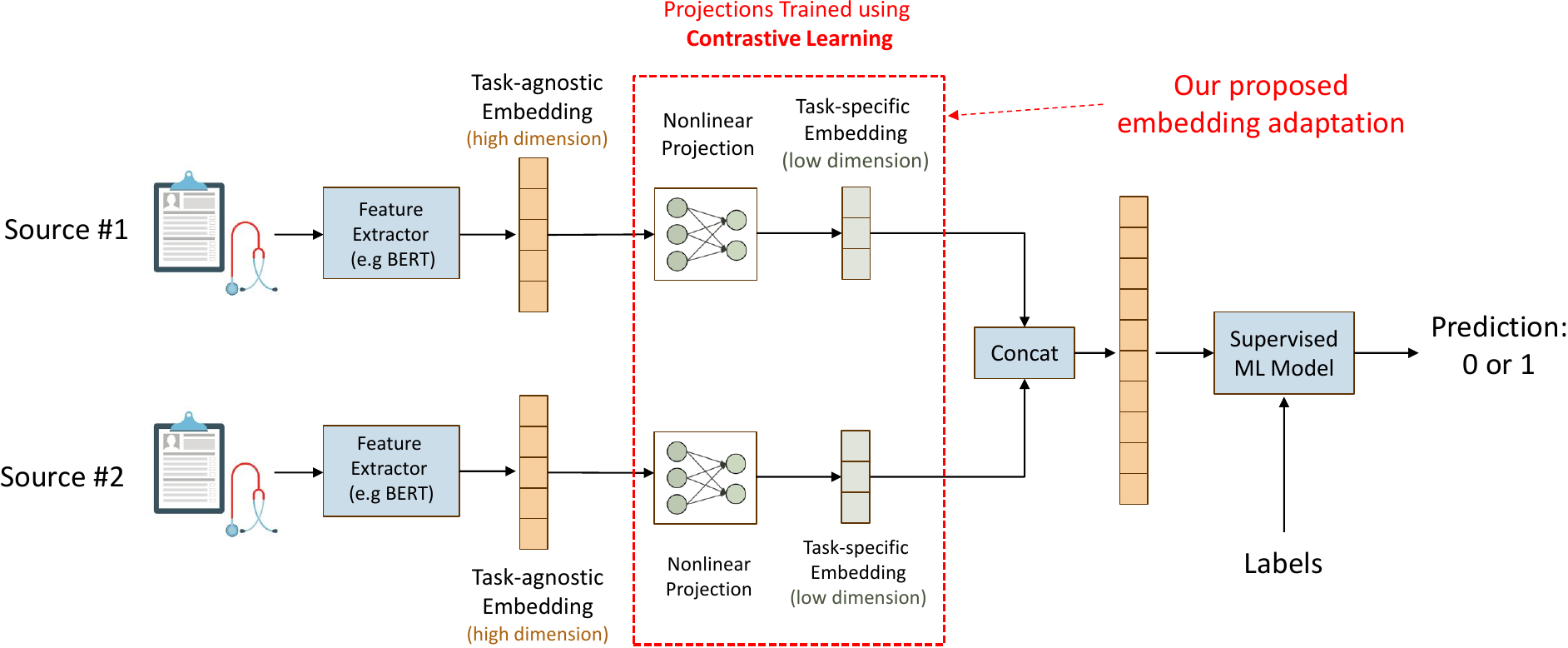}}
        \caption{Per-modality projection}
        \label{fig:per-modality}
    \end{subfigure}
    \caption{Multimodal prediction with task-specific embeddings}
    \label{fig:pipeline2}
    \end{center}
    \vskip -0.2in
\end{figure}

% Our approach adapts the embeddings to the downstream task through lightweight nonlinear projections which are trained through contrastive learning.\par 
Let $f(x): \mathbb{R}^M\rightarrow \mathbb{R}^K$ with $K<M$ be a dimension-reducing nonlinear projection.
We consider $2$ different projection paradigms which are shown in Figure (\ref{fig:pipeline2}):
\begin{itemize}
    \item \textbf{Single Projection}: The embeddings are concatenated into a single embedding $\bm{e}_i=[\bm{e_i}^{(1)}, \dots, \bm{e_i}^{(m)}]$ and the resulting vector is passed from the projection as $\bm{p}_i=f(\bm{e}_i)$.
    Then, we use the projected embedding $\bm{p}_i$ in our downstream task (Figure \ref{fig:single-proj}).
    %\newpage
    \item \textbf{Per-modality Projection}: Each embedding modality is projected using a different 
    projection ${\bm{p}_i^{(j)}=f_j(\bm{e}_i^{(j)}), \: j\in [m]}$, and the projections are then concatenated
    into a vector $\bm{p}=[\bm{p}_i^{(1)}, \dots, \bm{p}_i^{(m)}]$.
    This vector is used in our downstream task (Figure \ref{fig:per-modality}).
\end{itemize}
In both cases, we consider $f$ and $f_j$ to be small feed-forward ReLU NN projections,
and we use the labels $y_i$ to train these projections using Contrastive Learning.

To analyze the method, we will focus on the per-modality projection case,
since the single-projection case is a simpler extension of this. 
In that setting, let's focus on the projection of a single modality $\bm{p}_i^{(j)}=f_j(\bm{e}_i^{(j)})$.
To train our projection, we first get a random minibatch of size B from the original dataset,
which we assume without loss of generality that 
it contains the tuples $(\bm{e}_1^{(j)}, y_1), \dots (\bm{e}_B^{(j)}, y_B)$.
We then construct the following dataset of constrastive pairs:
\begin{equation}
    C_b^{(j)}= \{ \: (\bm{e}^{(j)}_i, \bm{e}^{(j)}_k, \mathds{1}\{y_i=y_k\})\:\}_{k\leq i,\:i\in [B]}
\end{equation}
For this batch, we calculate the following contrastive loss, which is similar to the SimCLR loss of \cite{chen_simple_2020}:
\begin{equation}
\begin{split}
        \mathcal{L}_b^{(j)}=-\sum_{(\bm{u}, \bm{v}, \ell)\in C_b^{(j)}}&\Big[\ell\cdot \log\sigma({g}^T_j(\bm{u}){g}_j(\bm{v})/\tau)+\\
        &+(1-\ell)\cdot \log\sigma({g}^T_j(\bm{u}){g}_j(\bm{v})/\tau)\Big].
\end{split}
\end{equation}
We repeat the process for the remaining minibatches of the original dataset, and we 
use this loss to train our nonlinear projection $f_j$.
If we are using per-modality projections, then each projection is trained independently from
the others.
If we are using a single projection, then the projection is trained exactly the same way,
but in this case, the input of $f$ is not a single-modality embedding $\bm{e}_j$, but rather
the concatenated embeddings $[{\bm{e}_i^{(1)}},\dots, {\bm{e}_i^{(m)}}]$.

Note that the loss $\mathcal{L}_b^{(j)}$ corresponds to a binary, cross-entropy loss with sigmoid activation.
Here our logit is ${g}^T_j(\bm{u}){g}_j(\bm{v})/\tau$
which is the similarity of the $2$ projections $\bm{u}$ and $\bm{v}$
divided by a temperature constant $\tau$.
Also, our label is $1$ iff both embeddings $\bm{u}$ and $\bm{v}$ correspond to the 
same label.
Intuitively, we want to learn a projection function $f_j$ such that
embeddings corresponding to the same label will have a similar projections
and embeddings corresponding to different labels will have different projections.\par 

%After training is 
% For every modality, we learn the projection function independently using standard back-propagation to minimize the nonlinear loss $\mathcal{L}^{(j)}$.
% Then, after the projections are learned, we project all of the modalities
% through their corresponding projections $g_j$ and we concatenate the
% resulting projections into a feature vector.
% This feature vector is then used for the classification task.
% This process can be seen in Figure (\ref{fig:pipeline2}). \par 

After training the projection(s), we use them to project our original embeddings and generate a task-specific embeddings for each data-point.
Then, we use these new embeddings, together with the labels $y_i$, to train a binary classification model.
This process is illustrated in Figure (\ref{fig:pipeline2}).

% Single column version

% Full width version (comment above and uncomment below)
% \begin{figure*}[t]
%     \vskip 0.2in
%     \begin{center}
%     \begin{subfigure}{0.48\textwidth}
%         \centerline{\includegraphics[width=\textwidth]{figs/cl_uni.png}}
%         \caption{Single Projection}
%         \label{fig:single-proj}
%     \end{subfigure}
%     \hfill
%     \begin{subfigure}{0.48\textwidth}
%         \centerline{\includegraphics[width=\textwidth]{figs/cl_multi.png}}
%         \caption{Per-modality projection}
%         \label{fig:per-modality}
%     \end{subfigure}
%     \caption{Multimodal prediction with task-specific embeddings}
%     \label{fig:pipeline2}
%     \end{center}
%     \vskip -0.2in
% \end{figure*}

\subsection{Datasets \& Feature Extractors}
% Due to the limited availability of public datasets that can be used for multimodal classification,
% we utilize $2$ datasets:
In order to evaluate our approach, we use $2$ different datasets: {A proprietary healthcare dataset from our ongoing collaboration with Hartford Healthcare Hospital System (HHC), and} a publicly-available multimodal movie dataset.
Below we describe these datasets and the ways that we preprocessed them.
We also outline the embedding-extractors we use for each dataset.
\subsubsection{Healthcare Data}
\label{sub:hhc}
{Through our IRB-approved research protocol, we have access to Electronic Health Records (EHRs) of 4000 patients from the EPIC database of HHC, which is a centralized clinical data registry that contains information on patient demographics, medical problems, laboratory data, and various clinical notes and documents}. 
From these records, we utilize {the following} distinct sources of data:
\begin{enumerate}
    \item The \textbf{History \& Physical} (\emph{H\&P}), which is a clinical note that contains information about {current and past conditions of the patient, risk factors, past medications, and other essential information that aids physicians to determine the individual's health status}.
    \item The \textbf{Progress Note} (\emph{Prg}), {which describes the patient's condition and treatment plan, and documents their progress or lack thereof between the previous and the most recent note}.
    \item The \textbf{ICD-9/ICD-10 diagnosis codes}, {which are standardized diagnostic codes used for clinical classification and billing purposes}.
\end{enumerate}
%\newpage
We treat these sources as $3$ independent data sources/modalities.
The first $2$ sources are textual, and can be passed directly from a language model to generate embeddings.
% For the 3rd source, we first convert the ICD9/ICD10 codes into a textual representation (following \cite{tabtext}).
For the 3rd source, we first convert the
International Classification of Diseases (ICD-9/ICD-10) codes into human-readable diagnoses texts using established mapping tables \cite{hcup}.
This process of converting structured, tabluar information into text
has been shown to yield better performance than using the data as structured features \cite{tabtext}.
% This process of converting tabular information to text is similar to \cite{tabtext}, and can be shown to improve performance than treating these data as structured features.
Then, we can also pass this text representation from a language model to extract embeddings.\par 
For each patient, we also have binary labels about the existence of various health conditions, such diabetes, hypertension, chronic lung disease, and others.
%These labels were collected by data managers for the National Database of the Society of Thoracic Surgeons (STS). \par 
Hence, given these labels, our supervised ML task is to predict the existence of the various conditions from the $3$ clinical sources we described.
In particular, in this work, we focus on predicting diabetes and hypertension from the EHR sources.\par 
For this dataset, we use $2$ different feature extractors to generate embeddings from the text sources.
We first use ClinicalBERT \cite{huang2019clinicalbert}, which has been used extensively for embedding extraction in Healthcare applications (e.g. see \cite{haim}). 
Secondly, we also wanted to try a more recent large-scale language model.
For this reason, we opted for Qwen2 \cite{qwen2} with 7B parameters, which is an open-source conversational model produced by Alibaba group.
However, since decoder-only models are note good for producing high-quality embeddings out of the box (i.e. due to their causal attention mask, \cite{llm2vec}), we use the GTE-pretrained version of Qwen2 (\cite{gte}), which adapts Qwen2 for embedding generation. 
Note that the reason we opted for using Qwen2 instead of Llama3, is because Qwen2 with GTE pretraining ranks among the top models in the Massive Text Embedding benchmark (MTEB, \cite{mteb}).

\subsubsection{Multimodal Movie Dataset}
\label{sub:imdb}
As our second dataset, we use \textit{mmimdb} \cite{mmimdb}, which is the largest publicly-available 
multimodal dataset for movie genre prediction.
From this dataset, we utilize the \textit{image}, \textit{title}, \textit{year}, \textit{rating} and the \textit{plot summary} of the movie
as our input data, which we use to predict whether a movie is a comedy or a drama (i.e. binary classification).
In particular, for each movie, we first generate a single piece of text which contains all the tabular information we mentioned.
An example is shown in Figure (\ref{fig:tabinfo}). 

\begin{figure}
%    \centering
    \begin{center}
        % \vskip 0.1in
        \begin{tcolorbox}[
            colback=yellow!10,
            colframe=black,
            sharp corners,
            boxrule=0.2mm
        ]
            \textbf{Title:} Titanic\\
            \textbf{Country:} United States\\
            \textbf{Plot:} \small A seventeen-year-old aristocrat, expecting to be married to a wealthy suitor, falls in love with a kind but poor artist aboard the luxurious, ill-fated R.M.S. Titanic.\\
            \textbf{Rating:} 7.9\\
            \textbf{Year:} 1997
        \end{tcolorbox}
        % \vskip 0.1in
        
    \end{center}
    \caption{Tabular movie information as a single text}
    \label{fig:tabinfo}
\end{figure}

Using this approach, we convert the movie's tabular information into text, 
and thus the data can now be parsed by an LLM, similar to \cite{tabtext}.
We also use the movie poster image as a seperate modality.
Hence, we now have a single text modality and a single image modality for our predictive task.\par 
Finally, as for the labels, we only keep movies that belong to either the Comedy genre or the Drama genre (those were the most populous classes in our dataset).
This way, we end up with a binary classification task with $3606$ datapoints.\par 
Finally, for the feature-extraction step, we use a different pretrained model
to generate embeddings from images and a different model to generate embeddings from text.
For images, we use a Vision Transformer (ViT-16, \cite{vit}) which is pretrained on the Imagenet 
dataset.
For text, we use a standard BERT model \cite{bert}.
Note that here we don't use the clinical version of BERT, since our application is non medical.

\section{Results}
In this section, we evaluate our approach in the $2$ datasets we described.
We perform this comparison for different feature extractors and downstream models, and we show that in the majority of cases, our projection method offers significant improvements compared to the other methods.

For our evaluation, we first use pretrained 
feature-extractors to generate task-agnostic embeddings for the different data modalities, as mentioned in Sections \ref{sub:hhc} and \ref{sub:imdb}.
% For our evaluation, we first use a pretrained feature-extractor to generate embeddings for the different data modalities.
Then, we test $3$ different approaches for performing supervised classification using the extracted embeddings:

\begin{itemize}[nosep, left=0pt]
    \item \textbf{Unprojected}: We fuse the embeddings from the different modalities together through simple concatenation.
    Then, we use a standard supervised ML model to predict the target label given the fused embeddings.
    \item \textbf{Contrastive}: We first project the embeddings into a lower-dimensional space 
    using the approach described in Section \ref{sec:approach}.
    Then, we use the resulting embeddings to perform supervised classification.
    \item \textbf{PCA}: We project the embeddings into a lower dimensional space using PCA,
    and then we use the projected  embeddings for classification.
\end{itemize}
For the projectiontion approaches (i.e. PCA and Constrastive projection), we experiment with
(i) Performing a single projection of the original, concatenated embeddings and (ii) Performing a different 
projection for each modality and concatenating the embeddings afterwards.
The $2$ approaches are illustrated in Figures (\ref{fig:single-proj}) and (\ref{fig:per-modality}) respectively.\par

For both PCA and the Constrastive projection, we use a projection size of $128$.
Then, in order to train the Contrastive projections, we use the hyperparameters shown in Table (\ref{tab:hyperparams}).\par

After projecting the embeddings with the different approaches, we 
use the embeddings in a binary classification task using  
RandomForest, XGBoost, CART, Logistic Regression, Support Vector Classifiers with RBF kernel and Multi-layer perceptrons with ReLU activations (MLP).
For all models, we use the default parameters and we don't perform hyperparameter tuning, in order to ensure fairness in our evaluation.
For comparing the different methods, we measure average out-of-sample F1 score using $5$-fold cross validation.
%over $5$ different splits of the data (i.e. we use $5$-fold cross validation).
We have also included detailed F1, ROC AUC and Accuracy results in Appendix \ref{sec:appendix}.
\par

%\subsection{Medical Dataset}
% We evaluate our method on predicting health-related variables, such as diabetes and hypertension
% from clinical notes.
% Our data comes from the national database of the Society of Thoraic Surgeons (STS) and the clinical records
% of Mass General Brigham (MGB).\par 
% We use data from $3$ text sources, including the History and Physical of the patient (H\&P), the Progress Notes and
% the text description of ICD9/ICD10 codes.
%We treat these sources as $3$ different modalities.\par 

%For this dataset, we have $3$ modalities and $9000$ patients, out of which we use $6000$ for training and $3000$ %for testing.
%For each of these patients, we first extract embeddings using ClinicalBERT and Qwen2 for the $3$ modalities.

\begin{table}[t]
    \caption{Contrastive learning hyperparameters}
    \label{tab:hyperparams}
    \vskip 0.15in
    \begin{center}
    \begin{small}
    \begin{sc}
    \begin{tabular}{lll}
    \toprule
    \textbf{Hyperparameter}     & \textbf{Medical} & \textbf{Movies} \\ \midrule
    Learning Rate ($\gamma$)              & $10^{-3} $              & $10^{-3}$              \\
    Minibatch Size (B)                 & 128                 & 128                 \\
    Number of Epochs (N)                      & 10                 & 15                \\ 
    Temperature ($\tau$)               & 0.1 & 0.1 \\
    Projection Hidden Layers            & 1 & 1\\
    Projection Size & 128 & 128\\
    \bottomrule
    \end{tabular}
    \end{sc}
    \end{small}
    \end{center}
    \vskip -0.1in
\end{table}

\begin{table}[h]
    \caption{Average F1 Score (\%) Across Experiments}
    \label{tab:avg_f1}
    %\vskip \skipamnt
    \begin{center}
    \begin{small}
    \begin{sc}
    \begin{threeparttable}
    \resizebox{1\columnwidth}{!}{%
    \begin{tabular}{l|l|l|l}
    \toprule
     Method & Diabetes & Hypertension & Movies \\
     %Method & & & \\
    %Method & F1 & F1 & F1 \\
        \midrule
        Unprojected & 58.3 {\tiny($\pm$23.5)} & 19.8 {\tiny($\pm$13.9)} & 56.2 {\tiny($\pm$27.3)} \\ 
        Single Proj (Contr.) & 59.9 {\tiny($\pm$27.6)} & 23.7 {\tiny($\pm$10.8)} & 64.7 {\tiny($\pm$24.8)} \\ 
        Single Proj (PCA) & 52.5 {\tiny($\pm$25.9)} & 17.1 {\tiny($\pm$10.9)} & 58.7 {\tiny($\pm$27.8)} \\ 
        Per-Mod (Contr.) & \textbf{76.3 {\tiny($\pm$5.4)}} & \textbf{27.6 {\tiny($\pm$6.3)}} & \textbf{76.4 {\tiny($\pm$2.4)}} \\ 
        Per-Mod (PCA) & 55.7 {\tiny($\pm$27.3)} & 20.6 {\tiny($\pm$13.4)} & 54.5 {\tiny($\pm$28.1)} \\  
        \bottomrule
    \end{tabular}
    }
    \begin{tablenotes}
        \scriptsize
        %\item Embeddings -- \textbf{Movies}: BERT+ViT; \textbf{Medical}: ClinicalBERT/Qwen2
        \item Values show mean F1 Scores \& standard deviations across 
        \item all models and feature extractors
    \end{tablenotes}    
    \end{threeparttable}
    \end{sc}
    \end{small}
    \end{center}
\end{table}
Table (\ref{tab:avg_f1}) shows the average performance of each projection method for each task (Diabetes, Hypertension, Movies).
We then proceed to analyze in detail the different tasks and the individual results.
%We then proceed to analyze the results for the different datasets.

\subsection{Medical Dataset}
For the medical dataset, we test the models on predicting diabetes and hypertension.
The results are shown
in Tables (\ref{tab:tab_diabetes}) and (\ref{tab:tab_hypertn}) respectively.

\begin{table}[t]
    \caption{$\Delta F1$ score (\%) for \emph{diabetes} with each projection method. Values show differences from unprojected baseline. Best results are shown in \textbf{bold}. }
    \label{tab:tab_diabetes}
    \vskip 0.15in
    \begin{center}
    \begin{small}
    \begin{sc}
    \begin{threeparttable}
    \resizebox{\columnwidth}{!}{%
    \begin{tabular}{ll|ll|ll}
    \toprule
    & & \multicolumn{2}{c|}{Single Proj.} & \multicolumn{2}{c}{Per-mod. Proj.} \\
    Extr. & Model & Contr. & PCA & Contr. & PCA \\
    \midrule
    % CBERT & CART & +20.1 {\tiny($\pm$1.9)} & -8.7 {\tiny($\pm$3.6)} & \textbf{+21.3} {\tiny($\pm$2.2)} & -4.2 {\tiny($\pm$1.6)} \\
    % CBERT & LR & -2.3 {\tiny($\pm$2.8)} & -2.4 {\tiny($\pm$1.4)} & \textbf{+2.0} {\tiny($\pm$1.5)} & +0.9 {\tiny($\pm$2.2)} \\
    % CBERT & MLP & -1.3 {\tiny($\pm$2.1)} & -5.7 {\tiny($\pm$1.6)} & \textbf{+2.8} {\tiny($\pm$1.6)} & +1.4 {\tiny($\pm$1.7)} \\
    % CBERT & RF & +14.2 {\tiny($\pm$2.5)} & -32.0 {\tiny($\pm$2.7)} & \textbf{+19.9} {\tiny($\pm$1.3)} & -37.6 {\tiny($\pm$4.3)} \\
    % CBERT & XGB & +2.1 {\tiny($\pm$2.7)} & -7.6 {\tiny($\pm$1.5)} & \textbf{+7.1} {\tiny($\pm$1.5)} & -1.6 {\tiny($\pm$1.3)} \\
    % \midrule
    % GQW & CART & +27.1 {\tiny($\pm$1.6)} & -5.1 {\tiny($\pm$2.8)} & \textbf{+30.5} {\tiny($\pm$2.5)} & +2.0 {\tiny($\pm$1.8)} \\
    % GQW & LR & +9.8 {\tiny($\pm$1.0)} & +4.3 {\tiny($\pm$2.4)} & \textbf{+15.2} {\tiny($\pm$2.1)} & +13.6 {\tiny($\pm$1.9)} \\
    % GQW & MLP & -2.6 {\tiny($\pm$1.3)} & -10.6 {\tiny($\pm$2.4)} & \textbf{+4.0} {\tiny($\pm$2.0)} & -0.5 {\tiny($\pm$2.2)} \\
    % GQW & RF & +43.3 {\tiny($\pm$2.3)} & -10.7 {\tiny($\pm$1.6)} & \textbf{+50.1} {\tiny($\pm$2.3)} & -2.0 {\tiny($\pm$3.6)} \\
    % GQW & XGB & +6.8 {\tiny($\pm$1.9)} & -13.8 {\tiny($\pm$0.6)} & \textbf{+13.4} {\tiny($\pm$1.8)} & +0.4 {\tiny($\pm$2.7)} \\
    CBERT &  CART & +18.2 {\tiny($\pm$3.2)} &  -7.3 {\tiny($\pm$1.9)} & \textbf{+18.3 {\tiny($\pm$2.0)}} &  -3.6 {\tiny($\pm$2.7)} \\
    CBERT &    LR &  -2.7 {\tiny($\pm$1.9)} &  -2.1 {\tiny($\pm$1.0)} &  \textbf{+0.1 {\tiny($\pm$1.9)}} &  -1.1 {\tiny($\pm$1.6)} \\
    CBERT &   MLP &  -2.0 {\tiny($\pm$1.6)} &  -4.2 {\tiny($\pm$1.7)} &  \textbf{+1.5 {\tiny($\pm$2.2)}} & +-0.1 {\tiny($\pm$1.3)} \\
    CBERT &   SVC &  +9.6 {\tiny($\pm$4.9)} &  +0.0 {\tiny($\pm$0.0)} & \textbf{+81.2 {\tiny($\pm$1.9)}} &  +0.0 {\tiny($\pm$0.0)} \\
    CBERT &   XGB &  +4.1 {\tiny($\pm$1.9)} &  -7.1 {\tiny($\pm$2.1)} &  \textbf{+7.0 {\tiny($\pm$2.6)}} &  -1.6 {\tiny($\pm$2.1)} \\
    \midrule
    GQW &  CART & +15.7 {\tiny($\pm$2.5)} &  -0.9 {\tiny($\pm$0.9)} & \textbf{+21.0 {\tiny($\pm$0.9)}} &  +1.3 {\tiny($\pm$2.0)} \\
    GQW &    LR &  +6.2 {\tiny($\pm$0.8)} &  +5.4 {\tiny($\pm$1.3)} & \textbf{+12.2 {\tiny($\pm$1.3)}} &  +9.9 {\tiny($\pm$1.4)} \\
    GQW &   MLP &  -3.5 {\tiny($\pm$1.3)} &  -5.9 {\tiny($\pm$1.7)} &  \textbf{+3.1 {\tiny($\pm$1.4)}} &  -1.2 {\tiny($\pm$1.5)} \\
    GQW &   SVC & -26.6 {\tiny($\pm$0.0)} & -26.6 {\tiny($\pm$0.0)} & \textbf{+43.9 {\tiny($\pm$0.9)}} & -26.6 {\tiny($\pm$0.0)} \\
    GQW &   XGB &  +6.2 {\tiny($\pm$2.2)} &  -6.9 {\tiny($\pm$1.3)} & \textbf{+12.9 {\tiny($\pm$1.8)}} &  +0.5 {\tiny($\pm$1.3)} \\
    \bottomrule
    \end{tabular}
    }
    \begin{tablenotes}
        \scriptsize  % Makes all notes smaller
        \item \makebox[0.6\columnwidth][l]{Extractors -- CBERT: ClinicalBERT, GQW: GTE+Qwen2}
        \item \makebox[0.6\columnwidth][l]{Models -- LR: Log. Regression, RF: Random Forest}
    \end{tablenotes}
    \end{threeparttable}
    \end{sc}
    \end{small}
    % \footnotesize{Extractors -- \textbf{CBERT}: ClinicalBERT, \textbf{GQW}: GTE+Qwen2. Models -- \textbf{LR}: Logistic Regression, \textbf{RF}: Random Forest, \textbf{XGB}: XGBoost.}
    \end{center}
    \vskip -0.1in
\end{table}

\begin{table}[t]
    \caption{$\Delta F1$ score (\%) for \emph{hypertension} with each projection method. Values show differences from unprojected baseline. Best results are shown in \textbf{bold}.}
    \label{tab:tab_hypertn}
    \vskip 0.15in
    \begin{center}
    \begin{small}
    \begin{sc}
    \begin{threeparttable}
    \resizebox{\columnwidth}{!}{%
    \begin{tabular}{ll|ll|ll}
    \toprule
    & & \multicolumn{2}{c|}{Single Proj.} & \multicolumn{2}{c}{Per-mod. Proj.} \\
    Extr. & Model & Contr. & PCA & Contr. & PCA \\
    \midrule
    CBERT &  CART &  \textbf{+3.6 {\tiny($\pm$4.4)}} &  -2.7 {\tiny($\pm$2.9)} &           +3.4 {\tiny($\pm$3.8)} &          +0.0 {\tiny($\pm$1.9)} \\
    CBERT &    LR &           -4.5 {\tiny($\pm$3.7)} &  -9.9 {\tiny($\pm$4.3)} &           -6.2 {\tiny($\pm$3.1)} & \textbf{+1.1 {\tiny($\pm$3.8)}} \\
    CBERT &   MLP &           -4.1 {\tiny($\pm$5.2)} &  -3.1 {\tiny($\pm$3.1)} &           -2.7 {\tiny($\pm$2.1)} & \textbf{+2.5 {\tiny($\pm$4.7)}} \\
    CBERT &   SVC &          +11.0 {\tiny($\pm$3.5)} &  +0.0 {\tiny($\pm$0.0)} & \textbf{+19.5 {\tiny($\pm$4.6)}} &          +0.0 {\tiny($\pm$0.0)} \\
    CBERT &   XGB &  \textbf{+9.3 {\tiny($\pm$4.3)}} &  -9.0 {\tiny($\pm$4.1)} &           +8.0 {\tiny($\pm$4.7)} &          -5.8 {\tiny($\pm$3.7)} \\
    \midrule
      GQW &  CART &           +7.0 {\tiny($\pm$2.7)} &  +4.0 {\tiny($\pm$1.9)} & \textbf{+13.0 {\tiny($\pm$5.9)}} &          +5.1 {\tiny($\pm$4.2)} \\
      GQW &    LR & \textbf{+26.5 {\tiny($\pm$4.1)}} & +10.9 {\tiny($\pm$3.3)} &          +21.8 {\tiny($\pm$2.7)} &         +24.3 {\tiny($\pm$2.9)} \\
      GQW &   MLP &           -1.7 {\tiny($\pm$3.7)} &  -2.3 {\tiny($\pm$3.1)} &           -7.3 {\tiny($\pm$5.9)} &          -1.8 {\tiny($\pm$3.9)} \\
      GQW &   SVC &           +0.0 {\tiny($\pm$0.0)} &  +0.0 {\tiny($\pm$0.0)} &  \textbf{+8.5 {\tiny($\pm$4.7)}} &          +0.0 {\tiny($\pm$0.0)} \\
      GQW &   XGB & \textbf{+19.9 {\tiny($\pm$3.5)}} &  -0.7 {\tiny($\pm$1.9)} &          +16.6 {\tiny($\pm$5.8)} &          +0.7 {\tiny($\pm$3.0)} \\
    \bottomrule
    \end{tabular}
    }
    \begin{tablenotes}
        \scriptsize  % Makes all notes smaller
        \item \makebox[0.6\columnwidth][l]{Extractors -- CBERT: ClinicalBERT, GQW: GTE+Qwen2}
        \item \makebox[0.6\columnwidth][l]{Models -- LR: Log. Regression, RF: Random Forest}
    \end{tablenotes}
    \end{threeparttable}
    \end{sc}
    \end{small}
    \end{center}
    \vskip -0.1in
\end{table}

% \begin{figure}[h]
%     \centering
%     \begin{subfigure}[t]{0.5\textwidth}
%         \centering
%         \includegraphics[height=2.6in]{figs/tsne1.png}
%         \caption{Original embedding}
%     \end{subfigure}%
%     ~ 
%     \begin{subfigure}[t]{0.5\textwidth}
%         \centering
%         \includegraphics[height=2.6in]{figs/tsne2.png}
%         \caption{Projected embedding (with Contrastive Learning)}
%     \end{subfigure}
%     \caption{Embedding visualization using t-SNE}
%     \label{fig:tsne}
% \end{figure}

% \begin{figure}[ht]
%     \vskip 0.2in
%     \begin{center}
%         \begin{subfigure}{\columnwidth}
%             \centerline{\includegraphics[width=\columnwidth]{figs/tsne1.png}}
%             \caption{Original embedding}
%         \end{subfigure}
%         \vskip 0.1in
%         \begin{subfigure}{\columnwidth}
%             \centerline{\includegraphics[width=\columnwidth]{figs/tsne2.png}}
%             \caption{Projected embedding (with Contrastive Learning)}
%         \end{subfigure}
%         \caption{Embedding visualization using t-SNE}
%         \label{fig:tsne}
%     \end{center}
%     \vskip -0.2in
% \end{figure}

\begin{figure}[ht]
    \vskip 0.2in
    \begin{center}
    \begin{subfigure}{0.48\columnwidth}
        \centerline{\includegraphics[width=\columnwidth]{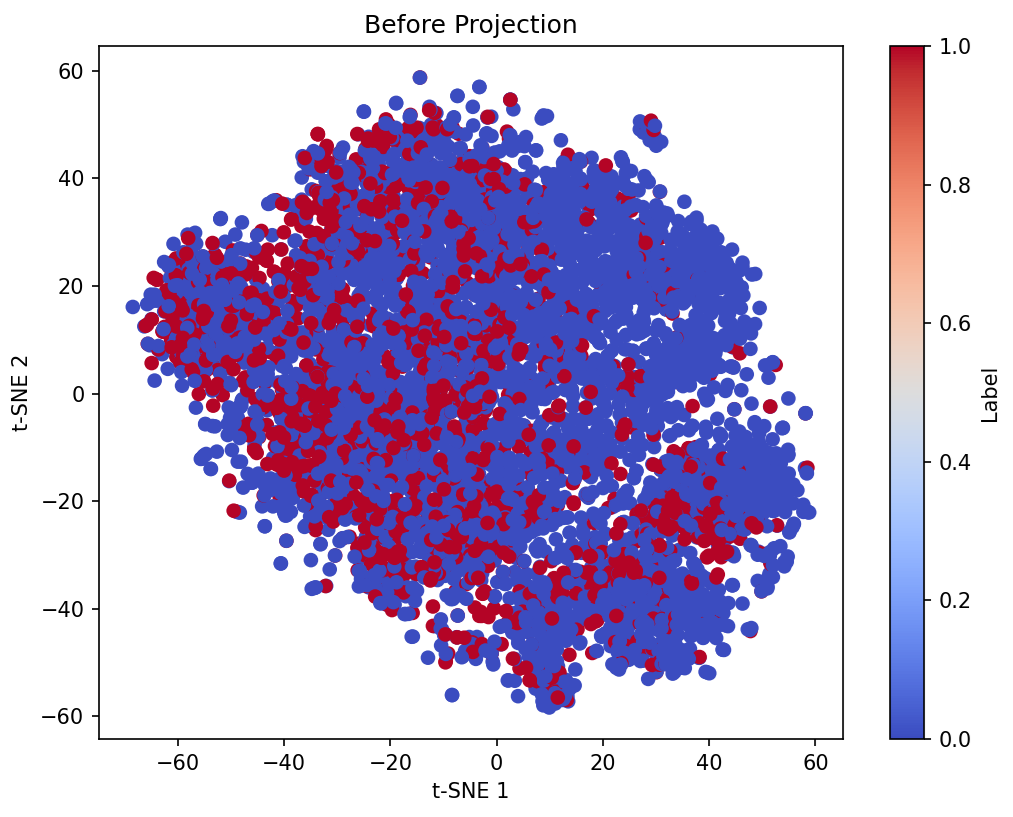}}
        \caption{Original embedding}
    \end{subfigure}
    \hfill
    \begin{subfigure}{0.48\columnwidth}
        \centerline{\includegraphics[width=\columnwidth]{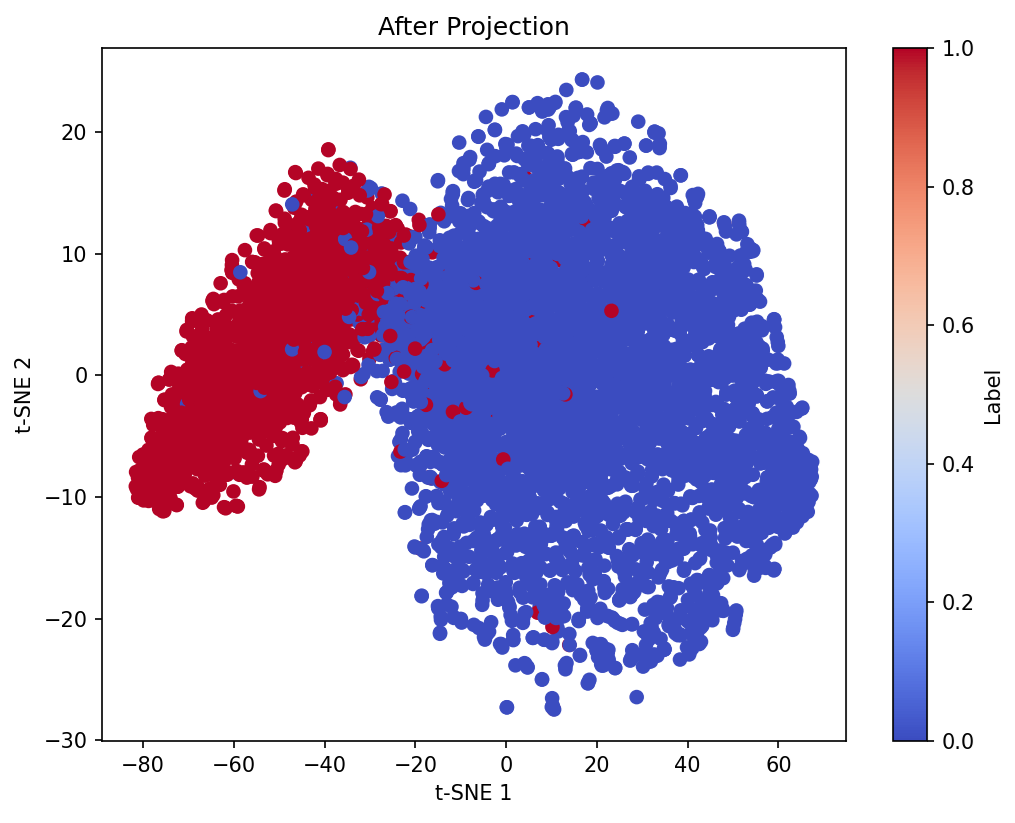}}
        \caption{Projected embedding}
    \end{subfigure}
    \caption{t-SNE plot of embeddings before and after contrastive projection.}
    \label{fig:tsne}
    \end{center}
    \vskip -0.2in
\end{figure}
%\newpage

%We test the method for predicting diabetes and hypertension from clinical notes.
%The results are shown in Table (\ref{tab:tab_diabetes}) and (\ref{tab:tab_hypertn}) respectively.\par 
Examining the results for predicting \textit{diabetes} (Table (\ref{tab:tab_diabetes})) and \textit{hypertension} (Table (\ref{tab:tab_hypertn})), we can see than in the vast majority of cases,
our contrastive projections beat 
both the unprojected embeddings and the embeddings projected using PCA.
In fact, for many combination of models and feature extractors, our method beats the baseline by more than $10\%$ in terms of F1 score,
demonstrating significant improvement at a very low computational cost.\par 

% Examining these results, we can clearly see that our Contrastive projection method almost universally beats
% both the unprojected embeddings and the embeddings projected using PCA.
% This picture seems to be consistent for different feature extractors (ClinicalBERT \& Qwen2) and different downstream models.
% In fact, in some cases, the Constrastive projection beats the unprojected baseline by more than $30\%$.
% In some cases, the difference between the contrastive projection 
To better understand this behavior, we focus on the task of predicting \textit{diabetes} using a single projection, and we perform a t-SNE visualization of the embeddings before and after the contrastive projection.
We also indicate the labels with a different color (i.e. blue indicates no diabetes and red indicates diabetes), as showin in Figure (\ref{fig:tsne}). 
From this figure, it is evident that before the projection, there doesn't seem to be a clear class separation, but after the contrastive projection, the label separation becomes very pronounced.
This explains why simpler ML models (i.e. models with simpler decision boundaries)
such as SVC and CART tend to perform much better when applied to the projected embeddings rather than the original embeddings.
% To better understand the effect of our method, we also perform a visualization of the fused embeddings using a t-SNE plot both before and after the constrastive projection.

% , as shown on Figure (\ref{fig:tsne}).
% From this figure, it is evident that before the projection, there doesn't seem to be a clear class separation, but after the contrastive projection, the label separation becomes very pronounced.

\subsection{Movie Dataset}
For the multimodal movie-genre prediction dataset, our downstream task was to predict whether the movie is comedy or drama.
The results are shown on Table (\ref{tab:tab_movie}).

\begin{table}[H]
    \caption{$\Delta F1$ score (\%) for \emph{movies} with each projection method. Values show differences from unprojected baseline. Best results are shown in \textbf{bold}.}
    \label{tab:tab_movie}
    \vskip 0.15in
    \begin{center}
    \begin{small}
    \begin{sc}
    \begin{threeparttable}
    \resizebox{\columnwidth}{!}{%
    \begin{tabular}{ll|ll|ll}
    \toprule
    & & \multicolumn{2}{c|}{Single Proj.} & \multicolumn{2}{c}{Per-mod. Proj.} \\
    Extr. & Model & Contr. & PCA & Contr. & PCA \\
    \midrule
    BViT &  CART &          +19.5 {\tiny($\pm$2.7)} &          +4.6 {\tiny($\pm$2.5)} & \textbf{+19.6 {\tiny($\pm$1.7)}} &          -2.5 {\tiny($\pm$1.7)} \\
     BViT &    LR &           -0.4 {\tiny($\pm$1.6)} & \textbf{+2.2 {\tiny($\pm$1.3)}} &           -0.7 {\tiny($\pm$2.4)} &          +1.7 {\tiny($\pm$1.9)} \\
     BViT &   MLP &           -2.0 {\tiny($\pm$1.0)} &          +0.6 {\tiny($\pm$1.2)} &           -1.5 {\tiny($\pm$3.3)} & \textbf{+0.8 {\tiny($\pm$1.9)}} \\
     BViT &    RF & \textbf{+20.3 {\tiny($\pm$2.3)}} &          +4.6 {\tiny($\pm$4.5)} &          +20.1 {\tiny($\pm$2.6)} &         -14.8 {\tiny($\pm$2.7)} \\
     BViT &   SVC &         +30.3 {\tiny($\pm$16.0)} &          +0.0 {\tiny($\pm$0.4)} & \textbf{+75.1 {\tiny($\pm$2.7)}} &          -0.3 {\tiny($\pm$0.0)} \\
     BViT &   XGB &           +3.4 {\tiny($\pm$2.5)} &          +2.8 {\tiny($\pm$2.2)} &  \textbf{+4.3 {\tiny($\pm$1.9)}} &          -0.8 {\tiny($\pm$1.8)} \\
    \bottomrule
    \end{tabular}
    }
    \begin{tablenotes}
        \scriptsize  % Makes all notes smaller
        \item \makebox[0.6\columnwidth][l]{Extractors -- BViT: BERT+ViT multimodal model}
        \item \makebox[0.6\columnwidth][l]{Models -- LR: Log. Regression, RF: Random Forest}
    \end{tablenotes}
    \end{threeparttable}
    \end{sc}
    \end{small}
    \end{center}
    \vskip -0.1in
\end{table}

We can see than in $3$ out of the $5$ cases, our constrastive projection outperforms the other methods by a 
significant margin.
These results are consistent with the medical dataset, showing that depending on the downstream model used,
the contrastive projection has the potential to offer significant performance improvements compared to conventional methods.

% can offer very significant performance benefits over unprojected embeddings,
% or embeddings projected with methods such as PCA.

\subsection{Time-Performance Trade-off}
In the previous sections, we established that our method can have significant performance benefits
in various downstream tasks.
In this section, we show that the computational overhead it introduces
is negligible, maintaining roughly the same execution time and resources as 
the unprojected baseline.
To further showcase the utility our method, we also compare the performance and resources
it uses against a scenario where the feature extractors of the different sources are
fully fine-tuned to the downstream task.
We then show that fine-tuning the feature extractors, despite having better downstream performance,
is orders of magnitude slower than our method, making it inaccessible in resource-constrained environments.
\par 
For the timing benchmarks, we focus on the task of predicting diabetes from the $3$ sources of the
Medical Dataset.
We use ClinicalBERT as the feature extractor for all sources, and we 
use MLP as our downstream model, so that the entire architecture (including the feature extractors)
can be fine-tuned using backpropagation.
We test $3$ different downstream training paradigms:
\begin{enumerate}
    \item \textbf{Unprojected Baseline}: Same as the baseline used in previous sections ( Figure (\ref{fig:pipeline1})), where MLP is the downstream model and feature-extractors are kept frozen.
    \item \textbf{Contrastive Per-modality projection}: Each modality is projected independently, and the projections are trained through Constrastive Learning (Figure (\ref{fig:per-modality})).
    MLP is used as the downstream model and feature-extractors are kept frozen.
    \item \textbf{Full Finetuning}: The initial embeddings are concatenated and a classification head is added on top of them. 
    Then, the entire architecture (including the ClinicalBERT feature extractors) is trained/finetuned to predict diabetes.
    A small number of epochs ($e=5$) is used to prevent catastrophic forgetting.
\end{enumerate}
For each of the methods, we run experiments (i) using only CPU and (ii) using both CPU and GPU.
The CPU and GPU used is an Intel(R) Xeon(R) Silver 4316 CPU @ 2.30GHz and an 80GB NVIDIA A100 GPU .
When using the CPU, we are only utilizing $8$ cores to emulate performance of a real-world resource-constrained scenario.\par
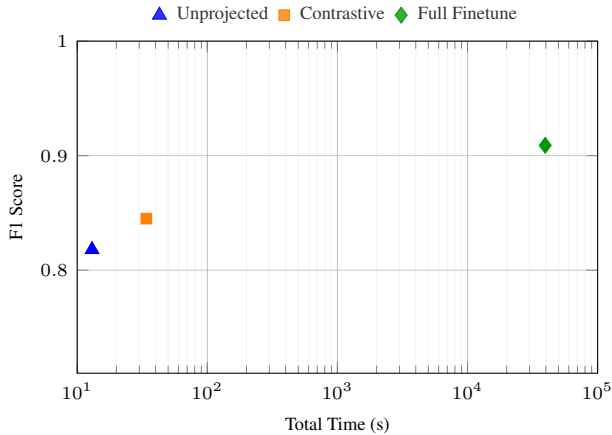
\begin{figure}
    \centering
    \pgfplotsset{compat=1.18}
\begin{tikzpicture}
    \begin{axis}[
        width=8.5cm, height=6cm,
        xlabel={Total Time (s)},
        ylabel={F1 Score},
        xmode=log,
        log basis x={10},
        ymin=0.71, ymax=1,
        xmin=10, xmax=100000,
        %legend style={at={(0.5,-0.3)}, anchor=north, legend columns=1, 
        %draw=none
        %},
        legend style={
            at={(0.5,1.02)},
            anchor=south,
            draw=white!100!black, % Semi-transparent border color
            fill opacity=0.8, % Adjust legend background opacity
            font=\scriptsize, % Smaller font size
            legend columns=3,
            column sep=0.2em
        },
        legend cell align={left},
        grid=both,
        minor grid style={opacity=0.1, color=gray},
        major grid style={opacity=0.4, color=gray},
        tick label style={font=\scriptsize},
        label style={font=\scriptsize},
    ]
         % Unprojected
        \addplot[
            only marks,
            mark=triangle*,
            mark size=3,
            color=blue
        ] coordinates {
            (13, 0.818)
        };
        \addlegendentry{Unprojected  }
        
        % Contrastive
        \addplot[
            only marks,
            mark=square*,
            color=orange,
        ] coordinates {
            (34, 0.845)
        };
        \addlegendentry{Contrastive  }
        
        % Full Finetune
        \addplot[
            only marks,
            mark=diamond*,
            mark size=3,
            color=green!60!black
        ] coordinates {
            (39562, 0.909)
        };
        \addlegendentry{Full Finetune  }
        
        % Annotations
        \node[anchor=west] at (axis cs:44,0.818) {};
        \node[anchor=west] at (axis cs:91,0.845) {};
        \node[anchor=west] at (axis cs:39562,0.909) {};
    \end{axis}
\end{tikzpicture}
    \caption{Time-performance trade-off}
    \label{fig:tradeoff}
\end{figure}

 
Due to high computational demands, and in order to ensure feairness, we run a single experiment from each type. 
In each of our experiments, we measure F1 score, training time and VRAM.
The results are reported in Table (\ref{tab:tradeoff}).
We also construct a time-performance trade-off plot (Figure (\ref{fig:tradeoff})).\par 
Comparing the different approaches, it is clear that our Contrastive approach has a very similar execution time as the unprojected baseline, both in the CPU and in the CPU+GPU scenario.
In fact, it is clear that our method doesn't even require a GPU, since it can run in less than a minute for $\sim$ 4k datapoints in a conventional CPU.
However, despite its low computational footprint, our method still offers a $3\%$ improvement in
diagnostic quality in terms of F1-score.\par 
On the other hand, fine-tuning the feature extractors does indeed offer a $6\%$ better performance than our method,
albeit at a much higher computational cost.
In fact, running the full fine-tuning procedure takes $3$ orders of more time (i.e. around $11$ hours)
and requires a powerful GPU with more than $32$ GB of VRAM, resources which are rarely found in
real-world hospital settings.
At the same time, it is evident that such fine-tuning methods are completely inapplicable in CPU-only environments.
\begin{table}[H]
    \caption{Performance \& Time of the different methods}
    \label{tab:tradeoff}
    \centering
    \begin{threeparttable}
    \begin{sc}
    \resizebox{\columnwidth}{!}{
    \begin{tabular}{lllll}
    \toprule
    \multirow{3}{*}{Method} & \multirow{3}{*}{F1} & \multicolumn{3}{c}{Resource} \\
    \cmidrule(lr){3-5}
    & & CPU & \multicolumn{2}{c}{GPU} \\
    \cmidrule(lr){3-3} \cmidrule(lr){4-5}
    & & Time (s) & Time (s) & VRAM (MB) \\
    \midrule
    Unprojected & 0.818 & 14 & 13 & -- \\
    Contrastive & 0.845 & 51 & 34 & 3 \\
    Full Finetune & 0.909 & -- & 39562 & 33770 \\
    % Unprojected & 0.797 & 14 & 13 & -- \\
    % Contrastive & 0.845 & 51 & 34 & 3 \\
    % Full Finetune & 0.909 & -- & 39562 & 33770 \\
    \bottomrule
    \end{tabular}
    }
     \begin{tablenotes}
      \small
      \item Time measured on 80GB NVIDIA A100 GPU /
      \item Intel(R) Xeon(R) Silver 4316 CPU @ 2.30GHz 
    \end{tablenotes}
    \end{sc}
    \end{threeparttable}
\end{table}

\section{Conclusion}
We introduced a general-purpose method that utilizes contrastive learning to learn a downstream task-specific non-linear projection to improve the performance of the prediction head. Our method runs in minutes, is low-resource, and offers significant improvements, without having to resort to expensive fine-tuning processes of foundation models.
Furthermore, it’s flexible, allowing for any multimodal combination, and can be applied for any binary classification downstream task. 
We showcase its effectiveness in healthcare, which is an ideal setting for our method; computational resources are scarce, patient data are ample, and performance is crucial.

\section*{Impact Statement}
This paper presents work whose goal is to advance the field of Machine Learning. 
There are many potential societal consequences of our work, none which we feel must be specifically highlighted here.

\clearpage
% Bibliography
\bibliography{ms}

\begin{thebibliography}{28}
\providecommand{\natexlab}[1]{#1}
\providecommand{\url}[1]{\texttt{#1}}
\expandafter\ifx\csname urlstyle\endcsname\relax
  \providecommand{\doi}[1]{doi: #1}\else
  \providecommand{\doi}{doi: \begingroup \urlstyle{rm}\Url}\fi

\bibitem[Arevalo et~al.(2017)Arevalo, Solorio, y~Gómez, and González]{mmimdb}
Arevalo, J., Solorio, T., y~Gómez, M.~M., and González, F.~A.
\newblock Gated multimodal units for information fusion, 2017.
\newblock URL \url{https://arxiv.org/abs/1702.01992}.

\bibitem[BehnamGhader et~al.(2024)BehnamGhader, Adlakha, Mosbach, Bahdanau,
  Chapados, and Reddy]{llm2vec}
BehnamGhader, P., Adlakha, V., Mosbach, M., Bahdanau, D., Chapados, N., and
  Reddy, S.
\newblock Llm2vec: Large language models are secretly powerful text encoders,
  2024.
\newblock URL \url{https://arxiv.org/abs/2404.05961}.

\bibitem[Brown(2020)]{brown2020language}
Brown, T.~B.
\newblock Language models are few-shot learners.
\newblock \emph{arXiv preprint arXiv:2005.14165}, 2020.

\bibitem[Bubeck et~al.(2023)Bubeck, Chandrasekaran, Eldan, Gehrke, Horvitz,
  Kamar, Lee, Lee, Li, Lundberg, et~al.]{bubeck2023sparks}
Bubeck, S., Chandrasekaran, V., Eldan, R., Gehrke, J., Horvitz, E., Kamar, E.,
  Lee, P., Lee, Y.~T., Li, Y., Lundberg, S., et~al.
\newblock Sparks of artificial general intelligence: Early experiments with
  gpt-4.
\newblock \emph{arXiv preprint arXiv:2303.12712}, 2023.

\bibitem[Carballo et~al.(2023)Carballo, Na, Ma, Boussioux, Zeng, Soenksen, and
  Bertsimas]{tabtext}
Carballo, K.~V., Na, L., Ma, Y., Boussioux, L., Zeng, C., Soenksen, L.~R., and
  Bertsimas, D.
\newblock Tabtext: A flexible and contextual approach to tabular data
  representation, 2023.
\newblock URL \url{https://arxiv.org/abs/2206.10381}.

\bibitem[Caron et~al.()Caron, Misra, Mairal, Goyal, Bojanowski, and
  Joulin]{caron_unsupervised_2020}
Caron, M., Misra, I., Mairal, J., Goyal, P., Bojanowski, P., and Joulin, A.
\newblock Unsupervised learning of visual features by contrasting cluster
  assignments.
\newblock In \emph{Advances in Neural Information Processing Systems},
  volume~33, pp.\  9912--9924. Curran Associates, Inc.
\newblock URL
  \url{https://papers.neurips.cc/paper_files/paper/2020/hash/70feb62b69f16e0238f741fab228fec2-Abstract.html}.

\bibitem[Chen et~al.()Chen, Kornblith, Norouzi, and Hinton]{chen_simple_2020}
Chen, T., Kornblith, S., Norouzi, M., and Hinton, G.
\newblock A simple framework for contrastive learning of visual
  representations.
\newblock URL \url{http://arxiv.org/abs/2002.05709}.

\bibitem[Devlin et~al.(2019)Devlin, Chang, Lee, and Toutanova]{bert}
Devlin, J., Chang, M.-W., Lee, K., and Toutanova, K.
\newblock Bert: Pre-training of deep bidirectional transformers for language
  understanding, 2019.
\newblock URL \url{https://arxiv.org/abs/1810.04805}.

\bibitem[Dosovitskiy(2020)]{dosovitskiy2020image}
Dosovitskiy, A.
\newblock An image is worth 16x16 words: Transformers for image recognition at
  scale.
\newblock \emph{arXiv preprint arXiv:2010.11929}, 2020.

\bibitem[Dosovitskiy et~al.(2021)Dosovitskiy, Beyer, Kolesnikov, Weissenborn,
  Zhai, Unterthiner, Dehghani, Minderer, Heigold, Gelly, Uszkoreit, and
  Houlsby]{vit}
Dosovitskiy, A., Beyer, L., Kolesnikov, A., Weissenborn, D., Zhai, X.,
  Unterthiner, T., Dehghani, M., Minderer, M., Heigold, G., Gelly, S.,
  Uszkoreit, J., and Houlsby, N.
\newblock An image is worth 16x16 words: Transformers for image recognition at
  scale, 2021.
\newblock URL \url{https://arxiv.org/abs/2010.11929}.

\bibitem[Dubey et~al.(2024)Dubey, Jauhri, Pandey, Kadian, Al-Dahle, Letman,
  Mathur, Schelten, Yang, Fan, Goyal, Hartshorn, Yang, Mitra, Sravankumar,
  Korenev, Hinsvark, Rao, Zhang, Rodriguez, Gregerson, Spataru, Roziere, Biron,
  Tang, Chern, Caucheteux, Nayak, Bi, Marra, McConnell, Keller, Touret, Wu,
  Wong, Ferrer, Nikolaidis, Allonsius, Song, Pintz, Livshits, Esiobu,
  Choudhary, Mahajan, Garcia-Olano, Perino, Hupkes, Lakomkin, AlBadawy,
  Lobanova, Dinan, Smith, Radenovic, Zhang, Synnaeve, Lee, Anderson, Nail,
  Mialon, Pang, Cucurell, Nguyen, Korevaar, Xu, Touvron, Zarov, Ibarra,
  Kloumann, Misra, Evtimov, Copet, Lee, Geffert, Vranes, Park, Mahadeokar,
  Shah, van~der Linde, Billock, Hong, Lee, Fu, Chi, Huang, Liu, Wang, Yu,
  Bitton, Spisak, Park, Rocca, Johnstun, Saxe, Jia, Alwala, Upasani, Plawiak,
  Li, Heafield, Stone, El-Arini, Iyer, Malik, Chiu, Bhalla, Rantala-Yeary,
  van~der Maaten, Chen, Tan, Jenkins, Martin, Madaan, Malo, Blecher, Landzaat,
  de~Oliveira, Muzzi, Pasupuleti, Singh, Paluri, Kardas, Oldham, Rita, Pavlova,
  Kambadur, Lewis, Si, Singh, Hassan, Goyal, Torabi, Bashlykov, Bogoychev,
  Chatterji, Duchenne, Çelebi, Alrassy, Zhang, Li, Vasic, Weng, Bhargava,
  Dubal, Krishnan, Koura, Xu, He, Dong, Srinivasan, Ganapathy, Calderer,
  Cabral, Stojnic, Raileanu, Girdhar, Patel, Sauvestre, Polidoro, Sumbaly,
  Taylor, Silva, Hou, Wang, Hosseini, Chennabasappa, Singh, Bell, Kim, Edunov,
  Nie, Narang, Raparthy, Shen, Wan, Bhosale, Zhang, Vandenhende, Batra,
  Whitman, Sootla, Collot, Gururangan, Borodinsky, Herman, Fowler, Sheasha,
  Georgiou, Scialom, Speckbacher, Mihaylov, Xiao, Karn, Goswami, Gupta,
  Ramanathan, Kerkez, Gonguet, Do, Vogeti, Petrovic, Chu, Xiong, Fu, Meers,
  Martinet, Wang, Tan, Xie, Jia, Wang, Goldschlag, Gaur, Babaei, Wen, Song,
  Zhang, Li, Mao, Coudert, Yan, Chen, Papakipos, Singh, Grattafiori, Jain,
  Kelsey, Shajnfeld, Gangidi, Victoria, Goldstand, Menon, Sharma, Boesenberg,
  Vaughan, Baevski, Feinstein, Kallet, Sangani, Yunus, Lupu, Alvarado, Caples,
  Gu, Ho, Poulton, Ryan, Ramchandani, Franco, Saraf, Chowdhury, Gabriel,
  Bharambe, Eisenman, Yazdan, James, Maurer, Leonhardi, Huang, Loyd, Paola,
  Paranjape, Liu, Wu, Ni, Hancock, Wasti, Spence, Stojkovic, Gamido, Montalvo,
  Parker, Burton, Mejia, Wang, Kim, Zhou, Hu, Chu, Cai, Tindal, Feichtenhofer,
  Civin, Beaty, Kreymer, Li, Wyatt, Adkins, Xu, Testuggine, David, Parikh,
  Liskovich, Foss, Wang, Le, Holland, Dowling, Jamil, Montgomery, Presani,
  Hahn, Wood, Brinkman, Arcaute, Dunbar, Smothers, Sun, Kreuk, Tian, Ozgenel,
  Caggioni, Guzmán, Kanayet, Seide, Florez, Schwarz, Badeer, Swee, Halpern,
  Thattai, Herman, Sizov, Guangyi, Zhang, Lakshminarayanan, Shojanazeri, Zou,
  Wang, Zha, Habeeb, Rudolph, Suk, Aspegren, Goldman, Damlaj, Molybog, Tufanov,
  Veliche, Gat, Weissman, Geboski, Kohli, Asher, Gaya, Marcus, Tang, Chan,
  Zhen, Reizenstein, Teboul, Zhong, Jin, Yang, Cummings, Carvill, Shepard,
  McPhie, Torres, Ginsburg, Wang, Wu, U, Saxena, Prasad, Khandelwal, Zand,
  Matosich, Veeraraghavan, Michelena, Li, Huang, Chawla, Lakhotia, Huang, Chen,
  Garg, A, Silva, Bell, Zhang, Guo, Yu, Moshkovich, Wehrstedt, Khabsa, Avalani,
  Bhatt, Tsimpoukelli, Mankus, Hasson, Lennie, Reso, Groshev, Naumov, Lathi,
  Keneally, Seltzer, Valko, Restrepo, Patel, Vyatskov, Samvelyan, Clark, Macey,
  Wang, Hermoso, Metanat, Rastegari, Bansal, Santhanam, Parks, White, Bawa,
  Singhal, Egebo, Usunier, Laptev, Dong, Zhang, Cheng, Chernoguz, Hart,
  Salpekar, Kalinli, Kent, Parekh, Saab, Balaji, Rittner, Bontrager, Roux,
  Dollar, Zvyagina, Ratanchandani, Yuvraj, Liang, Alao, Rodriguez, Ayub,
  Murthy, Nayani, Mitra, Li, Hogan, Battey, Wang, Maheswari, Howes, Rinott,
  Bondu, Datta, Chugh, Hunt, Dhillon, Sidorov, Pan, Verma, Yamamoto, Ramaswamy,
  Lindsay, Lindsay, Feng, Lin, Zha, Shankar, Zhang, Zhang, Wang, Agarwal,
  Sajuyigbe, Chintala, Max, Chen, Kehoe, Satterfield, Govindaprasad, Gupta,
  Cho, Virk, Subramanian, Choudhury, Goldman, Remez, Glaser, Best, Kohler,
  Robinson, Li, Zhang, Matthews, Chou, Shaked, Vontimitta, Ajayi, Montanez,
  Mohan, Kumar, Mangla, Albiero, Ionescu, Poenaru, Mihailescu, Ivanov, Li,
  Wang, Jiang, Bouaziz, Constable, Tang, Wang, Wu, Wang, Xia, Wu, Gao, Chen,
  Hu, Jia, Qi, Li, Zhang, Zhang, Adi, Nam, Yu, Wang, Hao, Qian, He, Rait,
  DeVito, Rosnbrick, Wen, Yang, and Zhao]{llama3}
Dubey, A., Jauhri, A., Pandey, A., Kadian, A., Al-Dahle, A., Letman, A.,
  Mathur, A., Schelten, A., Yang, A., Fan, A., Goyal, A., Hartshorn, A., Yang,
  A., Mitra, A., Sravankumar, A., Korenev, A., Hinsvark, A., Rao, A., Zhang,
  A., Rodriguez, A., Gregerson, A., Spataru, A., Roziere, B., Biron, B., Tang,
  B., Chern, B., Caucheteux, C., Nayak, C., Bi, C., Marra, C., McConnell, C.,
  Keller, C., Touret, C., Wu, C., Wong, C., Ferrer, C.~C., Nikolaidis, C.,
  Allonsius, D., Song, D., Pintz, D., Livshits, D., Esiobu, D., Choudhary, D.,
  Mahajan, D., Garcia-Olano, D., Perino, D., Hupkes, D., Lakomkin, E.,
  AlBadawy, E., Lobanova, E., Dinan, E., Smith, E.~M., Radenovic, F., Zhang,
  F., Synnaeve, G., Lee, G., Anderson, G.~L., Nail, G., Mialon, G., Pang, G.,
  Cucurell, G., Nguyen, H., Korevaar, H., Xu, H., Touvron, H., Zarov, I.,
  Ibarra, I.~A., Kloumann, I., Misra, I., Evtimov, I., Copet, J., Lee, J.,
  Geffert, J., Vranes, J., Park, J., Mahadeokar, J., Shah, J., van~der Linde,
  J., Billock, J., Hong, J., Lee, J., Fu, J., Chi, J., Huang, J., Liu, J.,
  Wang, J., Yu, J., Bitton, J., Spisak, J., Park, J., Rocca, J., Johnstun, J.,
  Saxe, J., Jia, J., Alwala, K.~V., Upasani, K., Plawiak, K., Li, K., Heafield,
  K., Stone, K., El-Arini, K., Iyer, K., Malik, K., Chiu, K., Bhalla, K.,
  Rantala-Yeary, L., van~der Maaten, L., Chen, L., Tan, L., Jenkins, L.,
  Martin, L., Madaan, L., Malo, L., Blecher, L., Landzaat, L., de~Oliveira, L.,
  Muzzi, M., Pasupuleti, M., Singh, M., Paluri, M., Kardas, M., Oldham, M.,
  Rita, M., Pavlova, M., Kambadur, M., Lewis, M., Si, M., Singh, M.~K., Hassan,
  M., Goyal, N., Torabi, N., Bashlykov, N., Bogoychev, N., Chatterji, N.,
  Duchenne, O., Çelebi, O., Alrassy, P., Zhang, P., Li, P., Vasic, P., Weng,
  P., Bhargava, P., Dubal, P., Krishnan, P., Koura, P.~S., Xu, P., He, Q.,
  Dong, Q., Srinivasan, R., Ganapathy, R., Calderer, R., Cabral, R.~S.,
  Stojnic, R., Raileanu, R., Girdhar, R., Patel, R., Sauvestre, R., Polidoro,
  R., Sumbaly, R., Taylor, R., Silva, R., Hou, R., Wang, R., Hosseini, S.,
  Chennabasappa, S., Singh, S., Bell, S., Kim, S.~S., Edunov, S., Nie, S.,
  Narang, S., Raparthy, S., Shen, S., Wan, S., Bhosale, S., Zhang, S.,
  Vandenhende, S., Batra, S., Whitman, S., Sootla, S., Collot, S., Gururangan,
  S., Borodinsky, S., Herman, T., Fowler, T., Sheasha, T., Georgiou, T.,
  Scialom, T., Speckbacher, T., Mihaylov, T., Xiao, T., Karn, U., Goswami, V.,
  Gupta, V., Ramanathan, V., Kerkez, V., Gonguet, V., Do, V., Vogeti, V.,
  Petrovic, V., Chu, W., Xiong, W., Fu, W., Meers, W., Martinet, X., Wang, X.,
  Tan, X.~E., Xie, X., Jia, X., Wang, X., Goldschlag, Y., Gaur, Y., Babaei, Y.,
  Wen, Y., Song, Y., Zhang, Y., Li, Y., Mao, Y., Coudert, Z.~D., Yan, Z., Chen,
  Z., Papakipos, Z., Singh, A., Grattafiori, A., Jain, A., Kelsey, A.,
  Shajnfeld, A., Gangidi, A., Victoria, A., Goldstand, A., Menon, A., Sharma,
  A., Boesenberg, A., Vaughan, A., Baevski, A., Feinstein, A., Kallet, A.,
  Sangani, A., Yunus, A., Lupu, A., Alvarado, A., Caples, A., Gu, A., Ho, A.,
  Poulton, A., Ryan, A., Ramchandani, A., Franco, A., Saraf, A., Chowdhury, A.,
  Gabriel, A., Bharambe, A., Eisenman, A., Yazdan, A., James, B., Maurer, B.,
  Leonhardi, B., Huang, B., Loyd, B., Paola, B.~D., Paranjape, B., Liu, B., Wu,
  B., Ni, B., Hancock, B., Wasti, B., Spence, B., Stojkovic, B., Gamido, B.,
  Montalvo, B., Parker, C., Burton, C., Mejia, C., Wang, C., Kim, C., Zhou, C.,
  Hu, C., Chu, C.-H., Cai, C., Tindal, C., Feichtenhofer, C., Civin, D., Beaty,
  D., Kreymer, D., Li, D., Wyatt, D., Adkins, D., Xu, D., Testuggine, D.,
  David, D., Parikh, D., Liskovich, D., Foss, D., Wang, D., Le, D., Holland,
  D., Dowling, E., Jamil, E., Montgomery, E., Presani, E., Hahn, E., Wood, E.,
  Brinkman, E., Arcaute, E., Dunbar, E., Smothers, E., Sun, F., Kreuk, F.,
  Tian, F., Ozgenel, F., Caggioni, F., Guzmán, F., Kanayet, F., Seide, F.,
  Florez, G.~M., Schwarz, G., Badeer, G., Swee, G., Halpern, G., Thattai, G.,
  Herman, G., Sizov, G., Guangyi, Zhang, Lakshminarayanan, G., Shojanazeri, H.,
  Zou, H., Wang, H., Zha, H., Habeeb, H., Rudolph, H., Suk, H., Aspegren, H.,
  Goldman, H., Damlaj, I., Molybog, I., Tufanov, I., Veliche, I.-E., Gat, I.,
  Weissman, J., Geboski, J., Kohli, J., Asher, J., Gaya, J.-B., Marcus, J.,
  Tang, J., Chan, J., Zhen, J., Reizenstein, J., Teboul, J., Zhong, J., Jin,
  J., Yang, J., Cummings, J., Carvill, J., Shepard, J., McPhie, J., Torres, J.,
  Ginsburg, J., Wang, J., Wu, K., U, K.~H., Saxena, K., Prasad, K., Khandelwal,
  K., Zand, K., Matosich, K., Veeraraghavan, K., Michelena, K., Li, K., Huang,
  K., Chawla, K., Lakhotia, K., Huang, K., Chen, L., Garg, L., A, L., Silva,
  L., Bell, L., Zhang, L., Guo, L., Yu, L., Moshkovich, L., Wehrstedt, L.,
  Khabsa, M., Avalani, M., Bhatt, M., Tsimpoukelli, M., Mankus, M., Hasson, M.,
  Lennie, M., Reso, M., Groshev, M., Naumov, M., Lathi, M., Keneally, M.,
  Seltzer, M.~L., Valko, M., Restrepo, M., Patel, M., Vyatskov, M., Samvelyan,
  M., Clark, M., Macey, M., Wang, M., Hermoso, M.~J., Metanat, M., Rastegari,
  M., Bansal, M., Santhanam, N., Parks, N., White, N., Bawa, N., Singhal, N.,
  Egebo, N., Usunier, N., Laptev, N.~P., Dong, N., Zhang, N., Cheng, N.,
  Chernoguz, O., Hart, O., Salpekar, O., Kalinli, O., Kent, P., Parekh, P.,
  Saab, P., Balaji, P., Rittner, P., Bontrager, P., Roux, P., Dollar, P.,
  Zvyagina, P., Ratanchandani, P., Yuvraj, P., Liang, Q., Alao, R., Rodriguez,
  R., Ayub, R., Murthy, R., Nayani, R., Mitra, R., Li, R., Hogan, R., Battey,
  R., Wang, R., Maheswari, R., Howes, R., Rinott, R., Bondu, S.~J., Datta, S.,
  Chugh, S., Hunt, S., Dhillon, S., Sidorov, S., Pan, S., Verma, S., Yamamoto,
  S., Ramaswamy, S., Lindsay, S., Lindsay, S., Feng, S., Lin, S., Zha, S.~C.,
  Shankar, S., Zhang, S., Zhang, S., Wang, S., Agarwal, S., Sajuyigbe, S.,
  Chintala, S., Max, S., Chen, S., Kehoe, S., Satterfield, S., Govindaprasad,
  S., Gupta, S., Cho, S., Virk, S., Subramanian, S., Choudhury, S., Goldman,
  S., Remez, T., Glaser, T., Best, T., Kohler, T., Robinson, T., Li, T., Zhang,
  T., Matthews, T., Chou, T., Shaked, T., Vontimitta, V., Ajayi, V., Montanez,
  V., Mohan, V., Kumar, V.~S., Mangla, V., Albiero, V., Ionescu, V., Poenaru,
  V., Mihailescu, V.~T., Ivanov, V., Li, W., Wang, W., Jiang, W., Bouaziz, W.,
  Constable, W., Tang, X., Wang, X., Wu, X., Wang, X., Xia, X., Wu, X., Gao,
  X., Chen, Y., Hu, Y., Jia, Y., Qi, Y., Li, Y., Zhang, Y., Zhang, Y., Adi, Y.,
  Nam, Y., Yu, Wang, Hao, Y., Qian, Y., He, Y., Rait, Z., DeVito, Z.,
  Rosnbrick, Z., Wen, Z., Yang, Z., and Zhao, Z.
\newblock The llama 3 herd of models, 2024.
\newblock URL \url{https://arxiv.org/abs/2407.21783}.

\bibitem[Gutmann \& Hyvärinen()Gutmann and
  Hyvärinen]{gutmann_noise-contrastive_2010}
Gutmann, M. and Hyvärinen, A.
\newblock Noise-contrastive estimation: A new estimation principle for
  unnormalized statistical models.
\newblock In \emph{Proceedings of the Thirteenth International Conference on
  Artificial Intelligence and Statistics}, pp.\  297--304. {JMLR} Workshop and
  Conference Proceedings.
\newblock URL \url{https://proceedings.mlr.press/v9/gutmann10a.html}.
\newblock {ISSN}: 1938-7228.

\bibitem[He et~al.()He, Fan, Wu, Xie, and Girshick]{he_momentum_2020}
He, K., Fan, H., Wu, Y., Xie, S., and Girshick, R.
\newblock Momentum contrast for unsupervised visual representation learning.
\newblock In \emph{2020 {IEEE}/{CVF} Conference on Computer Vision and Pattern
  Recognition ({CVPR})}, pp.\  9726--9735. {IEEE}.
\newblock ISBN 978-1-72817-168-5.
\newblock \doi{10.1109/CVPR42600.2020.00975}.
\newblock URL \url{https://ieeexplore.ieee.org/document/9157636/}.

\bibitem[{Healthcare Cost and Utilization Project (HCUP)}(2021)]{hcup}
{Healthcare Cost and Utilization Project (HCUP)}.
\newblock \emph{Introduction to the HCUP National Inpatient Sample (NIS)}.
\newblock Agency for Healthcare Research and Quality, Rockville, MD, Jun 2021.
\newblock URL
  \url{https://www.hcup-us.ahrq.gov/db/nation/nis/NIS_Introduction_2018.jsp}.
\newblock The National (Nationwide) Inpatient Sample database documentation.

\bibitem[Huang et~al.(2019)Huang, Altosaar, and
  Ranganath]{huang2019clinicalbert}
Huang, K., Altosaar, J., and Ranganath, R.
\newblock Clinicalbert: Modeling clinical notes and predicting hospital
  readmission.
\newblock \emph{arXiv preprint arXiv:1904.05342}, 2019.

\bibitem[Jia et~al.()Jia, Yang, Xia, Chen, Parekh, Pham, Le, Sung, Li, and
  Duerig]{jia_scaling_2021}
Jia, C., Yang, Y., Xia, Y., Chen, Y.-T., Parekh, Z., Pham, H., Le, Q.~V., Sung,
  Y., Li, Z., and Duerig, T.
\newblock Scaling up visual and vision-language representation learning with
  noisy text supervision.
\newblock URL \url{http://arxiv.org/abs/2102.05918}.

\bibitem[Li et~al.(2023)Li, Zhang, Zhang, Long, Xie, and Zhang]{gte}
Li, Z., Zhang, X., Zhang, Y., Long, D., Xie, P., and Zhang, M.
\newblock Towards general text embeddings with multi-stage contrastive
  learning, 2023.
\newblock URL \url{https://arxiv.org/abs/2308.03281}.

\bibitem[Mikolov(2013)]{mikolov2013efficient}
Mikolov, T.
\newblock Efficient estimation of word representations in vector space.
\newblock \emph{arXiv preprint arXiv:1301.3781}, 2013.

\bibitem[Muennighoff et~al.(2023)Muennighoff, Tazi, Magne, and Reimers]{mteb}
Muennighoff, N., Tazi, N., Magne, L., and Reimers, N.
\newblock Mteb: Massive text embedding benchmark, 2023.
\newblock URL \url{https://arxiv.org/abs/2210.07316}.

\bibitem[Oord et~al.()Oord, Li, and Vinyals]{oord_representation_2019}
Oord, A. v.~d., Li, Y., and Vinyals, O.
\newblock Representation learning with contrastive predictive coding.
\newblock URL \url{http://arxiv.org/abs/1807.03748}.

\bibitem[Oquab et~al.(2023)Oquab, Darcet, Moutakanni, Vo, Szafraniec, Khalidov,
  Fernandez, Haziza, Massa, El-Nouby, et~al.]{oquab2023dinov2}
Oquab, M., Darcet, T., Moutakanni, T., Vo, H., Szafraniec, M., Khalidov, V.,
  Fernandez, P., Haziza, D., Massa, F., El-Nouby, A., et~al.
\newblock Dinov2: Learning robust visual features without supervision.
\newblock \emph{arXiv preprint arXiv:2304.07193}, 2023.

\bibitem[Oquab et~al.(2024)Oquab, Darcet, Moutakanni, Vo, Szafraniec, Khalidov,
  Fernandez, Haziza, Massa, El-Nouby, Assran, Ballas, Galuba, Howes, Huang, Li,
  Misra, Rabbat, Sharma, Synnaeve, Xu, Jegou, Mairal, Labatut, Joulin, and
  Bojanowski]{dino}
Oquab, M., Darcet, T., Moutakanni, T., Vo, H., Szafraniec, M., Khalidov, V.,
  Fernandez, P., Haziza, D., Massa, F., El-Nouby, A., Assran, M., Ballas, N.,
  Galuba, W., Howes, R., Huang, P.-Y., Li, S.-W., Misra, I., Rabbat, M.,
  Sharma, V., Synnaeve, G., Xu, H., Jegou, H., Mairal, J., Labatut, P., Joulin,
  A., and Bojanowski, P.
\newblock Dinov2: Learning robust visual features without supervision, 2024.
\newblock URL \url{https://arxiv.org/abs/2304.07193}.

\bibitem[Radford et~al.()Radford, Kim, Hallacy, Ramesh, Goh, Agarwal, Sastry,
  Askell, Mishkin, Clark, Krueger, and Sutskever]{radford_learning_2021}
Radford, A., Kim, J.~W., Hallacy, C., Ramesh, A., Goh, G., Agarwal, S., Sastry,
  G., Askell, A., Mishkin, P., Clark, J., Krueger, G., and Sutskever, I.
\newblock Learning transferable visual models from natural language
  supervision.
\newblock URL \url{http://arxiv.org/abs/2103.00020}.

\bibitem[Radford et~al.(2021)Radford, Kim, Hallacy, Ramesh, Goh, Agarwal,
  Sastry, Askell, Mishkin, Clark, et~al.]{radford2021learning}
Radford, A., Kim, J.~W., Hallacy, C., Ramesh, A., Goh, G., Agarwal, S., Sastry,
  G., Askell, A., Mishkin, P., Clark, J., et~al.
\newblock Learning transferable visual models from natural language
  supervision.
\newblock In \emph{International conference on machine learning}, pp.\
  8748--8763. PMLR, 2021.

\bibitem[Soenksen et~al.(2022)Soenksen, Ma, Zeng, Boussioux,
  Villalobos~Carballo, Na, Wiberg, Li, Fuentes, and Bertsimas]{haim}
Soenksen, L.~R., Ma, Y., Zeng, C., Boussioux, L., Villalobos~Carballo, K., Na,
  L., Wiberg, H.~M., Li, M.~L., Fuentes, I., and Bertsimas, D.
\newblock Integrated multimodal artificial intelligence framework for
  healthcare applications.
\newblock \emph{npj Digital Medicine}, 5\penalty0 (1):\penalty0 149, Sep 2022.
\newblock ISSN 2398-6352.
\newblock \doi{10.1038/s41746-022-00689-4}.
\newblock URL \url{https://doi.org/10.1038/s41746-022-00689-4}.

\bibitem[Xia et~al.()Xia, Wilson, Goldstein, and Henao]{xiacontrastive}
Xia, M., Wilson, J., Goldstein, B., and Henao, R.
\newblock Contrastive learning for clinical outcome prediction with partial
  data sources.
\newblock In \emph{Forty-first International Conference on Machine Learning}.

\bibitem[Yang et~al.(2024)Yang, Yang, Hui, Zheng, Yu, Zhou, Li, Li, Liu, Huang,
  Dong, Wei, Lin, Tang, Wang, Yang, Tu, Zhang, Ma, Yang, Xu, Zhou, Bai, He,
  Lin, Dang, Lu, Chen, Yang, Li, Xue, Ni, Zhang, Wang, Peng, Men, Gao, Lin,
  Wang, Bai, Tan, Zhu, Li, Liu, Ge, Deng, Zhou, Ren, Zhang, Wei, Ren, Liu, Fan,
  Yao, Zhang, Wan, Chu, Liu, Cui, Zhang, Guo, and Fan]{qwen2}
Yang, A., Yang, B., Hui, B., Zheng, B., Yu, B., Zhou, C., Li, C., Li, C., Liu,
  D., Huang, F., Dong, G., Wei, H., Lin, H., Tang, J., Wang, J., Yang, J., Tu,
  J., Zhang, J., Ma, J., Yang, J., Xu, J., Zhou, J., Bai, J., He, J., Lin, J.,
  Dang, K., Lu, K., Chen, K., Yang, K., Li, M., Xue, M., Ni, N., Zhang, P.,
  Wang, P., Peng, R., Men, R., Gao, R., Lin, R., Wang, S., Bai, S., Tan, S.,
  Zhu, T., Li, T., Liu, T., Ge, W., Deng, X., Zhou, X., Ren, X., Zhang, X.,
  Wei, X., Ren, X., Liu, X., Fan, Y., Yao, Y., Zhang, Y., Wan, Y., Chu, Y.,
  Liu, Y., Cui, Z., Zhang, Z., Guo, Z., and Fan, Z.
\newblock Qwen2 technical report, 2024.
\newblock URL \url{https://arxiv.org/abs/2407.10671}.

\bibitem[Zhang et~al.()Zhang, Jiang, Miura, Manning, and
  Langlotz]{zhang_contrastive_2022}
Zhang, Y., Jiang, H., Miura, Y., Manning, C.~D., and Langlotz, C.~P.
\newblock Contrastive learning of medical visual representations from paired
  images and text.
\newblock URL \url{http://arxiv.org/abs/2010.00747}.

\end{thebibliography}
\bibliographystyle{icml2025}

% add appendix here
\clearpage
\appendix
% \onecolumn
\section{Additional Results}
\label{sec:appendix}

\subsection{Summary Results}
\label{apx:summary}

\newcommand{\skipamnt}{0.05in}

In Table (\ref{tab:avg_f1}), (\ref{tab:avg_auc}) and (\ref{tab:avg_accuracy}), we show the average F1, AUC and Accuracy scores across all models and feature extractors for each projection method and dataset.
Then, In table (\ref{tab:summary_f1}), (\ref{tab:summary_auc}) and (\ref{tab:summary_acc}), we show the best F1, AUC and Accuracy between all models and feature extractors for each projection method and task.

% \begin{table}[h]
%     \caption{Average F1 Score (\%) Across Experiments}
%     \label{tab:avg_f1}
%     \vskip \skipamnt
%     \begin{center}
%     \begin{small}
%     \begin{sc}
%     \begin{threeparttable}
%     \resizebox{1\columnwidth}{!}{%
%     \begin{tabular}{l|l|l|l}
%     \toprule
%      Method & Diabetes & Hypertension & Movies \\
%      %Method & & & \\
%     %Method & F1 & F1 & F1 \\
%         \midrule
%         Unprojected & 58.3 {\tiny($\pm$23.5)} & 19.8 {\tiny($\pm$13.9)} & 56.2 {\tiny($\pm$27.3)} \\ 
%         Single Proj (Contr.) & 59.9 {\tiny($\pm$27.6)} & 23.7 {\tiny($\pm$10.8)} & 64.7 {\tiny($\pm$24.8)} \\ 
%         Single Proj (PCA) & 52.5 {\tiny($\pm$25.9)} & 17.1 {\tiny($\pm$10.9)} & 58.7 {\tiny($\pm$27.8)} \\ 
%         Per-Mod (Contr.) & \textbf{76.3 {\tiny($\pm$5.4)}} & \textbf{27.6 {\tiny($\pm$6.3)}} & \textbf{76.4 {\tiny($\pm$2.4)}} \\ 
%         Per-Mod (PCA) & 55.7 {\tiny($\pm$27.3)} & 20.6 {\tiny($\pm$13.4)} & 54.5 {\tiny($\pm$28.1)} \\  
%         \bottomrule
%     \end{tabular}
%     }
%     \begin{tablenotes}
%         \scriptsize
%         %\item Embeddings -- \textbf{Movies}: BERT+ViT; \textbf{Medical}: ClinicalBERT/Qwen2
%         \item Values show mean F1 Scores \& standard deviations across 
%         \item all models and feature extractors
%     \end{tablenotes}    
%     \end{threeparttable}
%     \end{sc}
%     \end{small}
%     \end{center}
% \end{table}

\begin{table}[h]
    \caption{Average AUC Score (\%) Across Experiments}
    \label{tab:avg_auc}
    \vskip \skipamnt
    \begin{center}
    \begin{small}
    \begin{sc}
    \begin{threeparttable}
    \resizebox{1\columnwidth}{!}{%
    \begin{tabular}{l|l|l|l}
    \toprule
     Method & Diabetes & Hypertension & Movies \\
     %Method & & & \\
    %Method & F1 & F1 & F1 \\
        \midrule
        Unprojected & 78.0 {\tiny($\pm$15.0)} & 66.8 {\tiny($\pm$10.9)} & 79.6 {\tiny($\pm$16.5)} \\ 
        Single Proj (Contr.) & 79.1 {\tiny($\pm$14.2)} & 65.4 {\tiny($\pm$8.5)} & 82.4 {\tiny($\pm$14.2)} \\ 
        Single Proj (PCA) & 74.6 {\tiny($\pm$15.2)} & 64.1 {\tiny($\pm$9.3)} & 80.4 {\tiny($\pm$16.0)} \\ 
        Per-Mod (Contr.) & \textbf{86.9 {\tiny($\pm$5.8)}} & \textbf{68.5 {\tiny($\pm$8.5)}} & \textbf{87.7 {\tiny($\pm$5.1)}} \\ 
        Per-Mod (PCA) & 77.0 {\tiny($\pm$16.1)} & 65.2 {\tiny($\pm$9.9)} & 79.6 {\tiny($\pm$16.6)} \\   
        \bottomrule
    \end{tabular}
    }
    \begin{tablenotes}
        \scriptsize
        % \item Embeddings -- \textbf{Movies}: BERT+ViT; \textbf{Clinical Tasks}: ClinicalBERT
        % \item Values show mean AUC scores \& standard deviations across all models
        \item Values show mean AUC Scores \& standard deviations across 
        \item all models and feature extractors
    \end{tablenotes}    
    \end{threeparttable}
    \end{sc}
    \end{small}
    \end{center}
\end{table}

\begin{table}[h]
    \caption{Average Accuracy Score (\%) Across Experiments}
    \label{tab:avg_accuracy}
    \vskip \skipamnt
    \begin{center}
    \begin{small}
    \begin{sc}
    \begin{threeparttable}
    \resizebox{1\columnwidth}{!}{%
    \begin{tabular}{l|l|l|l}
    \toprule
     Method & Diabetes & Hypertension & Movies \\
     %Method & & & \\
    %Method & F1 & F1 & F1 \\
        \midrule
        Unprojected & 77.1 {\tiny($\pm$8.5)} & 83.2 {\tiny($\pm$3.5)} & 78.2 {\tiny($\pm$7.9)} \\ 
        Single Proj (Contr.) & 78.7 {\tiny($\pm$7.2)} & 83.1 {\tiny($\pm$1.7)} & 81.5 {\tiny($\pm$6.1)} \\ 
        Single Proj (PCA) & 74.3 {\tiny($\pm$8.2)} & 82.8 {\tiny($\pm$3.1)} & 79.5 {\tiny($\pm$7.5)} \\ 
        Per-Mod (Contr.) & \textbf{84.4 {\tiny($\pm$3.5)}} & \textbf{84.5 {\tiny($\pm$2.3)}} & \textbf{85.0 {\tiny($\pm$1.6)}} \\ 
        Per-Mod (PCA) & 76.3 {\tiny($\pm$8.7)} & 83.1 {\tiny($\pm$3.3)} & 77.9 {\tiny($\pm$8.1)} \\  
        \bottomrule
    \end{tabular}
    }
    \begin{tablenotes}
        \scriptsize
        % \item Embeddings -- \textbf{Movies}: BERT+ViT; \textbf{Clinical Tasks}: ClinicalBERT
        % \item Values show mean AUC scores \& standard deviations across all models
        \item Values show mean Accuracy Scores \& standard deviations across 
        \item all models and feature extractors
    \end{tablenotes}       
    \end{threeparttable}
    \end{sc}
    \end{small}
    \end{center}
\end{table}

\begin{table}[h]
    \caption{Best F1 Score (\%) Across Experiments}
    \label{tab:summary_f1}
    \vskip \skipamnt
    \begin{center}
    \begin{small}
    \begin{sc}
    \begin{threeparttable}
    \resizebox{1\columnwidth}{!}{%
    \begin{tabular}{l|cc|cc|cc}
    \toprule
     & \multicolumn{2}{c|}{Diabetes} & \multicolumn{2}{c|}{Hypertension} & \multicolumn{2}{c}{Movies} \\
    Method & F1 & Model & F1 & Model & F1 & Model \\
    \midrule
    Unprojected          &             81.2 &             LR &            35.5 &            LR &           77.9 &          MLP \\
Single Proj (Contr.) &             79.2 &             LR &            31.4 &           MLP &           76.9 &           LR \\
Single Proj (PCA)    &             79.3 &             LR &            29.3 &           MLP &  \textbf{79.2} &           LR \\
Per-Mod (Contr.)     &    \textbf{82.5} &             LR &            32.5 &           XGB &           77.4 &           LR \\
Per-Mod (PCA)        &             79.9 &             LR &   \textbf{38.4} &            LR &           79.2 &          MLP \\
    \bottomrule
    \end{tabular}
    }
    \begin{tablenotes}
        \scriptsize
        \item Models -- \textbf{LR}: Logistic Regression, \textbf{RF}: Random Forest
        \item Embeddings -- \textbf{Movies}: BERT+ViT; \textbf{Clinical Tasks}: ClinicalBERT
        % \item Values show mean Accuracy Scores \& standard deviations across 
        % \item all models and feature extractors
        \end{tablenotes}    
    \end{threeparttable}
    \end{sc}
    \end{small}
    \end{center}
\end{table}

\begin{table}[h]
    \caption{Best AUC Score (\%) Across Tasks}
    \label{tab:summary_auc}
    \vskip \skipamnt
    \begin{center}
    \begin{small}
    \begin{sc}
    \begin{threeparttable}
    \resizebox{1\columnwidth}{!}{%
    \begin{tabular}{l|cc|cc|cc}
    \toprule
     & \multicolumn{2}{c|}{Diabetes} & \multicolumn{2}{c|}{Hypertension} & \multicolumn{2}{c}{Movies} \\
    Method & AUC & Model & AUC & Model & AUC & Model \\
    \midrule
    Unprojected          &              92.9 &             LR &    \textbf{76.7} &            LR &            91.9 &           LR \\
    Single Proj (Contr.) &              91.4 &             LR &             73.4 &            LR &            91.4 &           LR \\
    Single Proj (PCA)    &              91.4 &             LR &             74.9 &            LR &   \textbf{92.8} &           LR \\
    Per-Mod (Contr.)     &     \textbf{93.7} &             LR &             76.3 &            LR &            91.6 &           LR \\
    Per-Mod (PCA)        &              92.5 &            MLP &             75.5 &            LR &            92.4 &          MLP \\
    \bottomrule
    \end{tabular}
    }
    \begin{tablenotes}
    \scriptsize
    \item Models -- \textbf{LR}: Logistic Regression, \textbf{RF}: Random Forest
    \item Embeddings -- \textbf{Movies}: BERT+ViT; \textbf{Clinical Tasks}: ClinicalBERT
    \end{tablenotes}    
    \end{threeparttable}
    \end{sc}
    \end{small}
    \end{center}
\end{table}

\begin{table}[H]
    \caption{Best Accuracy (\%) Across Experiments}
    \label{tab:summary_acc}
    \vskip \skipamnt
    \begin{center}
    \begin{small}
    \begin{sc}
    \begin{threeparttable}
    \resizebox{1\columnwidth}{!}{%
    \begin{tabular}{l|cc|cc|cc}
    \toprule
     & \multicolumn{2}{c|}{Diabetes} & \multicolumn{2}{c|}{Hypertension} & \multicolumn{2}{c}{Movies} \\
    Method & Acc & Model & Acc & Model & Acc & Model \\
    \midrule
    Unprojected          &                   87.5 &             LR &                  86.0 &           XGB &                 85.6 &          MLP \\
    Single Proj (Contr.) &                   86.1 &            XGB &                  85.2 &           SVC &                 84.9 &           LR \\
    Single Proj (PCA)    &                   86.3 &             LR &                  85.2 &            LR &        \textbf{86.7} &           LR \\
    Per-Mod (Contr.)     &          \textbf{88.4} &             LR &         \textbf{86.2} &           SVC &                 85.7 &           RF \\
    Per-Mod (PCA)        &                   86.6 &            MLP &                  85.2 &           XGB &                 86.5 &          MLP \\
    \bottomrule
    \end{tabular}
    }
    \begin{tablenotes}
    \setlength\labelsep{0pt}
    \scriptsize
    \item Models -- \textbf{LR}: Logistic Regression, \textbf{RF}: Random Forest
    \item Embeddings -- \textbf{Movies}: BERT+ViT; \textbf{Clinical Tasks}: ClinicalBERT
    \end{tablenotes}    
    \end{threeparttable}
    \end{sc}
    \end{small}
    \end{center}
\end{table}

\subsection{Detailed Breakdown of Unprojected Results}
\label{apx:unprojected}
In table (\ref{tab:unprojected_all}), we show the performance of the unprojected embeddings across all tasks and models.
This serves as the baseline for the other methods, from which we calculate the $\Delta$ of the performance of the other methods.

\begin{table}[h]
    \caption{Baseline (Unprojected) Performance Across All Tasks and Models}
    \label{tab:unprojected_all}
    \vskip \skipamnt
    \begin{center}
    \begin{small}
    \begin{sc}
    \begin{threeparttable}
    \resizebox{\columnwidth}{!}{%
    \begin{tabular}{lll|ccc}
    \toprule
    Task & Extractor & Model & F1 (\%) & AUC (\%) & Acc (\%) \\
    \midrule
    \multirow{10}{*}{\rotatebox[origin=c]{90}{Diabetes}} & \multirow{5}{*}{ClinicalBERT} & CART &  55.9 {\tiny ($\pm$2.2)} &  66.2 {\tiny ($\pm$1.5)} &  69.0 {\tiny ($\pm$1.1)} \\
           &       & LR &  81.2 {\tiny ($\pm$2.0)} &  92.9 {\tiny ($\pm$0.9)} &  87.5 {\tiny ($\pm$1.0)} \\
           &       & MLP &  79.7 {\tiny ($\pm$2.2)} &  91.6 {\tiny ($\pm$0.9)} &  86.4 {\tiny ($\pm$1.3)} \\
           &       & SVC &   0.1 {\tiny ($\pm$0.0)} &  50.0 {\tiny ($\pm$0.0)} &  65.4 {\tiny ($\pm$2.4)} \\
           &       & XGB &  74.4 {\tiny ($\pm$1.6)} &  90.5 {\tiny ($\pm$1.2)} &  84.4 {\tiny ($\pm$0.9)} \\
    \cmidrule{2-6}
           & \multirow{5}{*}{GTE+Qwen2} & CART &  45.9 {\tiny ($\pm$2.1)} &  58.7 {\tiny ($\pm$1.4)} &  62.5 {\tiny ($\pm$1.4)} \\
           &       & LR &  59.2 {\tiny ($\pm$1.6)} &  84.1 {\tiny ($\pm$0.5)} &  77.6 {\tiny ($\pm$1.4)} \\
           &       & MLP &  69.0 {\tiny ($\pm$1.2)} &  85.0 {\tiny ($\pm$0.4)} &  79.8 {\tiny ($\pm$0.8)} \\
           &       & SVC &  26.6 {\tiny ($\pm$4.6)} &  57.3 {\tiny ($\pm$1.6)} &  70.0 {\tiny ($\pm$3.1)} \\
           &       & XGB &  59.3 {\tiny ($\pm$1.4)} &  82.9 {\tiny ($\pm$0.7)} &  76.9 {\tiny ($\pm$2.3)} \\
    %\cline{1-6}
    %\cline{2-6}
    %\cmidrule{2-6}
    \midrule
    \multirow{10}{*}{\rotatebox[origin=c]{90}{Hypertension}} & \multirow{5}{*}{ClinicalBERT} & CART &  26.3 {\tiny ($\pm$3.3)} &  56.7 {\tiny ($\pm$2.1)} &  77.1 {\tiny ($\pm$1.6)} \\
           &       & LR &  35.5 {\tiny ($\pm$1.8)} &  76.7 {\tiny ($\pm$2.5)} &  85.0 {\tiny ($\pm$0.9)} \\
           &       & MLP &  35.0 {\tiny ($\pm$4.2)} &  74.3 {\tiny ($\pm$2.6)} &  83.0 {\tiny ($\pm$1.1)} \\
           &       & SVC &   0.1 {\tiny ($\pm$0.0)} &  50.0 {\tiny ($\pm$0.0)} &  85.1 {\tiny ($\pm$1.3)} \\
           &       & XGB &  21.1 {\tiny ($\pm$4.8)} &  75.3 {\tiny ($\pm$2.8)} &  86.0 {\tiny ($\pm$1.4)} \\
    \cmidrule{2-6}
           & \multirow{5}{*}{GTE+Qwen2} & CART &  17.2 {\tiny ($\pm$3.7)} &  51.4 {\tiny ($\pm$1.8)} &  75.3 {\tiny ($\pm$1.0)} \\
           &       & LR &   2.7 {\tiny ($\pm$2.2)} &  74.3 {\tiny ($\pm$1.2)} &  85.2 {\tiny ($\pm$1.4)} \\
           &       & MLP &  28.9 {\tiny ($\pm$4.1)} &  73.5 {\tiny ($\pm$1.4)} &  84.3 {\tiny ($\pm$1.5)} \\
           &       & SVC &   0.1 {\tiny ($\pm$0.0)} &  50.0 {\tiny ($\pm$0.0)} &  85.1 {\tiny ($\pm$1.3)} \\
           &       & XGB &   6.9 {\tiny ($\pm$2.4)} &  71.5 {\tiny ($\pm$1.0)} &  85.0 {\tiny ($\pm$1.1)} \\
    %\cline{1-6}
    %\cmidrule{2-6}
    \midrule
    \multirow{6}{*}{\rotatebox[origin=c]{90}{Movies}} & \multirow{6}{*}{bert+ViT} & CART &  53.5 {\tiny ($\pm$3.4)} &  65.1 {\tiny ($\pm$2.1)} &  68.8 {\tiny ($\pm$1.8)} \\
           &       & LR &  77.4 {\tiny ($\pm$0.3)} &  91.9 {\tiny ($\pm$0.4)} &  85.3 {\tiny ($\pm$0.5)} \\
           &       & MLP &  77.9 {\tiny ($\pm$2.9)} &  91.9 {\tiny ($\pm$0.4)} &  85.6 {\tiny ($\pm$1.8)} \\
           &       & RF &  56.4 {\tiny ($\pm$3.2)} &  87.8 {\tiny ($\pm$0.8)} &  78.8 {\tiny ($\pm$1.3)} \\
           &       & SVC &   0.3 {\tiny ($\pm$0.4)} &  50.1 {\tiny ($\pm$0.1)} &  66.9 {\tiny ($\pm$1.3)} \\
           &       & XGB &  72.1 {\tiny ($\pm$1.5)} &  90.5 {\tiny ($\pm$0.7)} &  83.5 {\tiny ($\pm$0.6)} \\
    \bottomrule
    \end{tabular}
    }
    \begin{tablenotes}
    \scriptsize
    \item All metrics are in percentages (\%)
    \item LR: Logistic Regression, RF: Random Forest, XGB: XGBoost
    \item Standard deviations shown in parentheses
    \end{tablenotes}
    \end{threeparttable}
    \end{sc}
    \end{small}
    \end{center}
    \vskip -0.1in
\end{table}

\subsection{Detailed Breakdown of Results}
\label{apx:breakdown}

In this section, we provide detailed breakdowns of model performance using additional metrics beyond F1 score. 
For each dataset, we show the $\Delta$ of both accuracy (tables \ref{tab:diabetes_acc}, \ref{tab:hypertn_acc}, \ref{tab:movies_acc}) and AUC (tables \ref{tab:diabetes_auc}, \ref{tab:hypertn_auc}, \ref{tab:movies_auc}) improvements over the baseline (unprojected) method, which is reported in table (\ref{tab:unprojected_all}).

% \subsubsection{Accuracy Results}
% \label{apx:accuracy}

% Table (\ref{tab:diabetes_acc}), (\ref{tab:hypertn_acc}) and (\ref{tab:movies_acc}) show the $\Delta$ accuracy results for the diabetes, hypertension, and movies datasets respectively.
%\clearpage
\begin{table}[t]
    \caption{$\Delta$ Accuracy (\%) for \emph{diabetes} with each projection method. Values show differences from unprojected baseline. Best results are shown in \textbf{bold}.}
    \label{tab:diabetes_acc}
    \vskip 0.15in
    \begin{center}
    \begin{small}
    \begin{sc}
    \begin{threeparttable}
    \resizebox{\columnwidth}{!}{%
    \begin{tabular}{ll|ll|ll}
    \toprule
    & & \multicolumn{2}{c|}{Single Proj.} & \multicolumn{2}{c}{Per-mod. Proj.} \\
    Extr. & Model & Contr. & PCA & Contr. & PCA \\
    \midrule
    CBERT &  CART & \textbf{+13.4 {\tiny($\pm$1.9)}} & -4.8 {\tiny($\pm$2.4)} &          +13.0 {\tiny($\pm$0.9)} & -1.6 {\tiny($\pm$1.8)} \\
    CBERT &    LR &           -1.9 {\tiny($\pm$0.7)} & -1.4 {\tiny($\pm$1.1)} &  \textbf{+0.1 {\tiny($\pm$1.2)}} & -1.1 {\tiny($\pm$1.2)} \\
    CBERT &   MLP &           -1.1 {\tiny($\pm$1.0)} & -2.7 {\tiny($\pm$1.1)} &  \textbf{+1.2 {\tiny($\pm$1.4)}} & +0.1 {\tiny($\pm$0.7)} \\
    CBERT &   SVC &           +1.7 {\tiny($\pm$2.8)} & +0.0 {\tiny($\pm$2.4)} & \textbf{+22.6 {\tiny($\pm$1.0)}} & +0.0 {\tiny($\pm$2.4)} \\
    CBERT &   XGB &           +1.5 {\tiny($\pm$0.5)} & -4.1 {\tiny($\pm$1.8)} &  \textbf{+3.3 {\tiny($\pm$1.4)}} & -0.6 {\tiny($\pm$1.0)} \\
    \midrule
    GQW &  CART &          +11.4 {\tiny($\pm$1.3)} & +0.1 {\tiny($\pm$1.9)} & \textbf{+14.7 {\tiny($\pm$1.4)}} & +1.2 {\tiny($\pm$2.2)} \\
    GQW &    LR &           -0.6 {\tiny($\pm$1.1)} & +0.7 {\tiny($\pm$1.8)} &  \textbf{+3.8 {\tiny($\pm$1.0)}} & +2.3 {\tiny($\pm$0.9)} \\
    GQW &   MLP &           -2.2 {\tiny($\pm$1.5)} & -4.3 {\tiny($\pm$1.6)} &  \textbf{+2.0 {\tiny($\pm$1.0)}} & -1.0 {\tiny($\pm$1.3)} \\
    GQW &   SVC &           -4.7 {\tiny($\pm$2.4)} & -4.7 {\tiny($\pm$2.4)} & \textbf{+11.2 {\tiny($\pm$0.9)}} & -4.7 {\tiny($\pm$2.4)} \\
    GQW &   XGB &           +0.5 {\tiny($\pm$1.4)} & -3.4 {\tiny($\pm$1.6)} &  \textbf{+4.6 {\tiny($\pm$1.5)}} & +0.1 {\tiny($\pm$1.0)} \\
    \bottomrule
    \end{tabular}
    }
    \begin{tablenotes}
        \scriptsize  % Makes all notes smaller
        \item \makebox[0.6\columnwidth][l]{Extractors -- CBERT: ClinicalBERT, GQW: GTE+Qwen2}
        \item \makebox[0.6\columnwidth][l]{Models -- LR: Log. Regression, RF: Random Forest}
    \end{tablenotes}
    \end{threeparttable}
    \end{sc}
    \end{small}
    \end{center}
    \vskip -0.1in
\end{table}

\begin{table}[t]
    \caption{$\Delta$ Accuracy (\%) for \emph{hypertension} with each projection method. Values show differences from unprojected baseline. Best results are shown in \textbf{bold}.}
    \label{tab:hypertn_acc}
    \vskip 0.15in
    \begin{center}
    \begin{small}
    \begin{sc}
    \begin{threeparttable}
    \resizebox{\columnwidth}{!}{%
    \begin{tabular}{ll|ll|ll}
    \toprule
    & & \multicolumn{2}{c|}{Single Proj.} & \multicolumn{2}{c}{Per-mod. Proj.} \\
    Extr. & Model & Contr. & PCA & Contr. & PCA \\
    \midrule
    CBERT &  CART & \textbf{+2.8 {\tiny($\pm$0.6)}} &         +-0.0 {\tiny($\pm$1.2)} &          +1.2 {\tiny($\pm$1.5)} &          +0.1 {\tiny($\pm$0.7)} \\
    CBERT &    LR &          -2.1 {\tiny($\pm$0.9)} &          +0.1 {\tiny($\pm$0.9)} & \textbf{+0.9 {\tiny($\pm$0.9)}} &          -0.5 {\tiny($\pm$1.2)} \\
    CBERT &   MLP &          +0.2 {\tiny($\pm$0.9)} &          +0.5 {\tiny($\pm$0.8)} &          +2.0 {\tiny($\pm$0.9)} & \textbf{+2.1 {\tiny($\pm$1.3)}} \\
    CBERT &   SVC &         +-0.1 {\tiny($\pm$1.1)} &          +0.0 {\tiny($\pm$1.3)} & \textbf{+0.7 {\tiny($\pm$1.3)}} &          +0.0 {\tiny($\pm$1.3)} \\
    CBERT &   XGB &          -2.6 {\tiny($\pm$0.7)} &          -1.0 {\tiny($\pm$1.2)} &          -1.6 {\tiny($\pm$0.9)} &          -0.5 {\tiny($\pm$0.9)} \\
    \midrule
    GQW &  CART &          +4.3 {\tiny($\pm$1.7)} &          +3.6 {\tiny($\pm$1.5)} & \textbf{+7.4 {\tiny($\pm$0.7)}} &          +2.8 {\tiny($\pm$1.3)} \\
    GQW &    LR &          -2.5 {\tiny($\pm$1.1)} &          -0.3 {\tiny($\pm$1.0)} &          -0.2 {\tiny($\pm$1.2)} &          -1.6 {\tiny($\pm$1.4)} \\
    GQW &   MLP &          -1.0 {\tiny($\pm$1.1)} &          -2.3 {\tiny($\pm$1.4)} & \textbf{+1.2 {\tiny($\pm$1.0)}} &          -1.5 {\tiny($\pm$1.2)} \\
    GQW &   SVC & \textbf{+0.0 {\tiny($\pm$1.3)}} & \textbf{+0.0 {\tiny($\pm$1.3)}} &          -0.2 {\tiny($\pm$0.8)} & \textbf{+0.0 {\tiny($\pm$1.3)}} \\
    GQW &   XGB &          -1.6 {\tiny($\pm$1.1)} &         +-0.1 {\tiny($\pm$1.5)} &          -0.2 {\tiny($\pm$0.8)} &         +-0.0 {\tiny($\pm$1.2)} \\
    \bottomrule
    \end{tabular}
    }
    \begin{tablenotes}
        \scriptsize  % Makes all notes smaller
        \item \makebox[0.6\columnwidth][l]{Extractors -- CBERT: ClinicalBERT, GQW: GTE+Qwen2}
        \item \makebox[0.6\columnwidth][l]{Models -- LR: Log. Regression, RF: Random Forest}
    \end{tablenotes}
    \end{threeparttable}
    \end{sc}
    \end{small}
    \end{center}
    \vskip -0.1in
\end{table}

\begin{table}[t]
    \caption{$\Delta$ Accuracy (\%) for \emph{movies} with each projection method. Values show differences from unprojected baseline. Best results are shown in \textbf{bold}.}
    \label{tab:movies_acc}
    \vskip 0.15in
    \begin{center}
    \begin{small}
    \begin{sc}
    \begin{threeparttable}
    \resizebox{\columnwidth}{!}{%
    \begin{tabular}{ll|ll|ll}
    \toprule
    & & \multicolumn{2}{c|}{Single Proj.} & \multicolumn{2}{c}{Per-mod. Proj.} \\
    Extr. & Model & Contr. & PCA & Contr. & PCA \\
    \midrule
     BViT &  CART &         +13.2 {\tiny($\pm$1.4)} &          +3.3 {\tiny($\pm$1.1)} & \textbf{+13.7 {\tiny($\pm$0.9)}} &  -0.5 {\tiny($\pm$0.9)} \\
     BViT &    LR &          -0.3 {\tiny($\pm$1.1)} & \textbf{+1.6 {\tiny($\pm$0.8)}} &           -0.2 {\tiny($\pm$1.6)} &  +1.2 {\tiny($\pm$0.9)} \\
     BViT &   MLP &          -1.1 {\tiny($\pm$0.4)} & \textbf{+0.5 {\tiny($\pm$0.9)}} &           -0.8 {\tiny($\pm$2.2)} &  +0.5 {\tiny($\pm$1.3)} \\
     BViT &    RF & \textbf{+6.0 {\tiny($\pm$1.1)}} &          +1.3 {\tiny($\pm$1.9)} &           +5.9 {\tiny($\pm$1.7)} &  -4.1 {\tiny($\pm$1.3)} \\
     BViT &   SVC &          +5.1 {\tiny($\pm$3.6)} &         +-0.0 {\tiny($\pm$1.3)} & \textbf{+17.5 {\tiny($\pm$1.8)}} & +-0.1 {\tiny($\pm$1.3)} \\
     BViT &   XGB &          +0.7 {\tiny($\pm$1.3)} &          +0.8 {\tiny($\pm$0.8)} &  \textbf{+1.3 {\tiny($\pm$1.3)}} &  -0.6 {\tiny($\pm$1.3)} \\
    \bottomrule
    \end{tabular}
    }
    \begin{tablenotes}
        \scriptsize  % Makes all notes smaller
        \item \makebox[0.6\columnwidth][l]{Extractors -- BViT: BERT+ViT multimodal model}
        \item \makebox[0.6\columnwidth][l]{Models -- LR: Log. Regression, RF: Random Forest}
    \end{tablenotes}
    \end{threeparttable}
    \end{sc}
    \end{small}
    \end{center}
    \vskip -0.1in
\end{table}

% \subsubsection{AUC Results}
% \label{apx:auc}
% Table (\ref{tab:diabetes_auc}), (\ref{tab:hypertn_auc}) and (\ref{tab:movies_auc}) show the $\Delta$ AUC results for the diabetes, hypertension, and movies datasets respectively.

\begin{table}[t]
    \caption{$\Delta$ AUC (\%) for \emph{diabetes} with each projection method. Values show differences from unprojected baseline. Best results are shown in \textbf{bold}.}
    \label{tab:diabetes_auc}
    \vskip 0.15in
    \begin{center}
    \begin{small}
    \begin{sc}
    \begin{threeparttable}
    \resizebox{\columnwidth}{!}{%
    \begin{tabular}{ll|ll|ll}
    \toprule
    & & \multicolumn{2}{c|}{Single Proj.} & \multicolumn{2}{c}{Per-mod. Proj.} \\
    Extr. & Model & Contr. & PCA & Contr. & PCA \\
    \midrule
    CBERT &  CART & +13.9 {\tiny($\pm$1.8)} & -5.5 {\tiny($\pm$2.2)} & \textbf{+14.2 {\tiny($\pm$1.3)}} & -2.5 {\tiny($\pm$2.1)} \\
    CBERT &    LR &  -1.6 {\tiny($\pm$1.0)} & -1.3 {\tiny($\pm$0.7)} &  \textbf{+0.3 {\tiny($\pm$0.6)}} & -0.9 {\tiny($\pm$1.3)} \\
    CBERT &   MLP &  -1.4 {\tiny($\pm$0.6)} & -2.1 {\tiny($\pm$0.9)} &  \textbf{+0.4 {\tiny($\pm$1.0)}} & +0.3 {\tiny($\pm$1.2)} \\
    CBERT &   SVC &  +2.5 {\tiny($\pm$1.4)} & +0.0 {\tiny($\pm$0.0)} & \textbf{+35.0 {\tiny($\pm$1.4)}} & +0.0 {\tiny($\pm$0.0)} \\
    CBERT &   XGB &  +0.4 {\tiny($\pm$1.4)} & -4.1 {\tiny($\pm$1.6)} &  \textbf{+1.8 {\tiny($\pm$0.9)}} & -0.8 {\tiny($\pm$0.7)} \\
    \midrule
    GQW &  CART & +12.0 {\tiny($\pm$1.3)} & -0.4 {\tiny($\pm$1.2)} & \textbf{+15.9 {\tiny($\pm$1.1)}} & +1.1 {\tiny($\pm$1.9)} \\
    GQW &    LR &  -0.8 {\tiny($\pm$1.1)} & -0.4 {\tiny($\pm$0.9)} &  \textbf{+2.8 {\tiny($\pm$0.3)}} & +1.6 {\tiny($\pm$0.8)} \\
    GQW &   MLP &  -2.5 {\tiny($\pm$1.2)} & -5.3 {\tiny($\pm$1.4)} &  \textbf{+2.2 {\tiny($\pm$0.6)}} & -1.0 {\tiny($\pm$0.7)} \\
    GQW &   SVC &  -7.3 {\tiny($\pm$0.0)} & -7.3 {\tiny($\pm$0.0)} & \textbf{+20.1 {\tiny($\pm$0.6)}} & -7.3 {\tiny($\pm$0.0)} \\
    GQW &   XGB &  -0.8 {\tiny($\pm$0.6)} & -6.3 {\tiny($\pm$1.6)} &  \textbf{+3.5 {\tiny($\pm$0.5)}} & +0.1 {\tiny($\pm$1.5)} \\
    \bottomrule
    \end{tabular}
    }
    \begin{tablenotes}
        \scriptsize  % Makes all notes smaller
        \item \makebox[0.6\columnwidth][l]{Extractors -- CBERT: ClinicalBERT, GQW: GTE+Qwen2}
        \item \makebox[0.6\columnwidth][l]{Models -- LR: Log. Regression, RF: Random Forest}
    \end{tablenotes}
    \end{threeparttable}
    \end{sc}
    \end{small}
    \end{center}
    \vskip -0.1in
 \end{table}
 
 \begin{table}[t]
    \caption{$\Delta$ AUC (\%) for \emph{hypertension} with each projection method. Values show differences from unprojected baseline. Best results are shown in \textbf{bold}.}
    \label{tab:hypertn_auc}
    \vskip 0.15in
    \begin{center}
    \begin{small}
    \begin{sc}
    \begin{threeparttable}
    \resizebox{\columnwidth}{!}{%
    \begin{tabular}{ll|ll|ll}
    \toprule
    & & \multicolumn{2}{c|}{Single Proj.} & \multicolumn{2}{c}{Per-mod. Proj.} \\
    Extr. & Model & Contr. & PCA & Contr. & PCA \\
    \midrule
    CBERT &  CART & \textbf{+2.4 {\tiny($\pm$2.4)}} & -1.5 {\tiny($\pm$1.9)} &          +2.0 {\tiny($\pm$1.9)} &          +0.2 {\tiny($\pm$1.4)} \\
    CBERT &    LR &          -4.0 {\tiny($\pm$4.0)} & -1.8 {\tiny($\pm$2.9)} &          -1.1 {\tiny($\pm$2.9)} &          -0.5 {\tiny($\pm$1.7)} \\
    CBERT &   MLP &          -3.6 {\tiny($\pm$2.8)} & -2.5 {\tiny($\pm$3.3)} &          -1.0 {\tiny($\pm$2.3)} & \textbf{+0.8 {\tiny($\pm$1.1)}} \\
    CBERT &   SVC &          +2.6 {\tiny($\pm$0.9)} & +0.0 {\tiny($\pm$0.0)} & \textbf{+5.2 {\tiny($\pm$1.5)}} &          +0.0 {\tiny($\pm$0.0)} \\
    CBERT &   XGB &          -6.0 {\tiny($\pm$3.3)} & -3.8 {\tiny($\pm$2.4)} &          -2.0 {\tiny($\pm$2.4)} &         +-0.1 {\tiny($\pm$2.8)} \\
    \midrule
      GQW &  CART &          +4.5 {\tiny($\pm$1.5)} & +2.9 {\tiny($\pm$1.2)} & \textbf{+7.9 {\tiny($\pm$3.1)}} &          +3.3 {\tiny($\pm$2.3)} \\
      GQW &    LR &          -4.0 {\tiny($\pm$1.6)} & -1.1 {\tiny($\pm$1.5)} &         +-0.1 {\tiny($\pm$1.8)} &          -2.9 {\tiny($\pm$2.1)} \\
      GQW &   MLP &          -3.1 {\tiny($\pm$1.7)} & -6.1 {\tiny($\pm$1.8)} & \textbf{+2.4 {\tiny($\pm$2.3)}} &          -3.9 {\tiny($\pm$2.2)} \\
      GQW &   SVC &          +0.0 {\tiny($\pm$0.0)} & +0.0 {\tiny($\pm$0.0)} & \textbf{+1.9 {\tiny($\pm$1.2)}} &          +0.0 {\tiny($\pm$0.0)} \\
      GQW &   XGB &          -3.1 {\tiny($\pm$3.0)} & -3.3 {\tiny($\pm$2.1)} & \textbf{+0.5 {\tiny($\pm$2.1)}} &          -1.6 {\tiny($\pm$1.1)} \\
    \bottomrule
    \end{tabular}
    }
    \begin{tablenotes}
        \scriptsize  % Makes all notes smaller
        \item \makebox[0.6\columnwidth][l]{Extractors -- CBERT: ClinicalBERT, GQW: GTE+Qwen2}
        \item \makebox[0.6\columnwidth][l]{Models -- LR: Log. Regression, RF: Random Forest}
    \end{tablenotes}
    \end{threeparttable}
    \end{sc}
    \end{small}
    \end{center}
    \vskip -0.1in
 \end{table}
 
 \begin{table}[t]
    \caption{$\Delta$ AUC (\%) for \emph{movies} with each projection method. Values show differences from unprojected baseline. Best results are shown in \textbf{bold}.}
    \label{tab:movies_auc}
    \vskip 0.15in
    \begin{center}
    \begin{small}
    \begin{sc}
    \begin{threeparttable}
    \resizebox{\columnwidth}{!}{%
    \begin{tabular}{ll|ll|ll}
    \toprule
    & & \multicolumn{2}{c|}{Single Proj.} & \multicolumn{2}{c}{Per-mod. Proj.} \\
    Extr. & Model & Contr. & PCA & Contr. & PCA \\
    \midrule
     BViT &  CART & \textbf{+14.7 {\tiny($\pm$1.7)}} &          +3.5 {\tiny($\pm$1.5)} &          +14.6 {\tiny($\pm$1.3)} &          -1.4 {\tiny($\pm$1.0)} \\
     BViT &    LR &           -0.5 {\tiny($\pm$0.5)} & \textbf{+0.9 {\tiny($\pm$0.3)}} &           -0.9 {\tiny($\pm$0.9)} &          +0.3 {\tiny($\pm$0.4)} \\
     BViT &   MLP &           -1.1 {\tiny($\pm$0.8)} &          +0.2 {\tiny($\pm$0.2)} &           -1.3 {\tiny($\pm$1.3)} & \textbf{+0.6 {\tiny($\pm$0.7)}} \\
     BViT &    RF &           +1.9 {\tiny($\pm$0.6)} &          -0.7 {\tiny($\pm$1.0)} &  \textbf{+2.4 {\tiny($\pm$0.9)}} &          -1.1 {\tiny($\pm$0.4)} \\
     BViT &   SVC &           +8.9 {\tiny($\pm$5.3)} &         +-0.0 {\tiny($\pm$0.1)} & \textbf{+31.3 {\tiny($\pm$2.3)}} &         +-0.1 {\tiny($\pm$0.0)} \\
     BViT &   XGB &          +-0.0 {\tiny($\pm$0.9)} &         +-0.1 {\tiny($\pm$0.9)} &  \textbf{+0.4 {\tiny($\pm$0.9)}} &          -0.8 {\tiny($\pm$0.5)} \\
    \bottomrule
    \end{tabular}
    }
    \begin{tablenotes}
        \scriptsize  % Makes all notes smaller
        \item \makebox[0.6\columnwidth][l]{Extractors -- BViT: BERT+ViT multimodal model}
        \item \makebox[0.6\columnwidth][l]{Models -- LR: Log. Regression, RF: Random Forest}
    \end{tablenotes}
    \end{threeparttable}
    \end{sc}
    \end{small}
    \end{center}
    \vskip -0.1in
 \end{table}

\end{document}